\documentclass[journal]{IEEEtran}

\usepackage{amsmath}
\usepackage{amssymb}
\usepackage{graphicx}

\usepackage{subfigure}
\usepackage[ruled]{algorithm2e}

\begin{document}

\title{\LARGE Multi-Agent Reinforcement Learning for Unmanned Aerial Vehicle Coordination by Multi-Critic Policy Gradient Optimization}

\author{Yoav~Alon and Huiyu Zhou

\thanks{Y. Alon and H. Zhou are with the School of Informatics, University of Leicester, United Kingdom.
E-mail: \{ya89;hz143\}@leicester.ac.uk}.}%

\markboth{IEEE Transactions on Robotics,~submitted in Dec. 2020}%
{Shell \MakeLowercase{\textit{et al.}}: Bare Demo of IEEEtran.cls for IEEE Journals}

\maketitle

\begin{abstract}

Recent technological progress in the development of Unmanned Aerial Vehicles (UAVs) together with decreasing acquisition costs make the application of drone fleets attractive for a wide variety of tasks. In agriculture, disaster management, search and rescue operations, commercial and military applications, the advantage of applying a fleet of drones originates from their ability to cooperate autonomously. Multi-Agent Reinforcement Learning approaches that aim to optimize a neural network based control policy, such as the best performing actor-critic policy gradient algorithms, struggle to effectively back-propagate errors of distinct rewards signal sources and tend to favor lucrative signals while neglecting coordination and exploitation of previously learned similarities. We propose a Multi-Critic Policy Optimization architecture with multiple value estimating networks and a novel advantage function that optimizes a stochastic actor policy network to achieve optimal coordination of agents. Consequently, we apply the algorithm to several tasks that require the collaboration of multiple drones in a physics-based reinforcement learning environment. Our approach achieves a stable policy network update and similarity in reward signal development for an increasing number of agents. The resulting policy achieves optimal coordination and compliance with constraints such as collision avoidance.


\end{abstract}


\IEEEpeerreviewmaketitle
\section{Introduction}

In recent years, technological progress in the development of Unmanned Aerial Vehicles (UAVs) together with decreasing acquisition costs makes the application of groups of drones attractive for a wide variety of areas. In agriculture, disaster management, search and rescue operations, commercial and military applications, the advantage of applying a group of drones originates from their ability to cooperate autonomously. The ability to plan and coordinate enables drones to share work load and increase efficiency for their aspired tasks. At the same time, certain safety constraints such as collision avoidance have to be ensured. Rather than using handcrafted policies for motion control, one should approach a continuous control task such as automated control of a group of UAVs with Multi-Agent Reinforcement Learning (MARL) \cite{MARL, DeepMARL, MultiAgentRL, PartiallyMARL, disertationCooperationMARL}.

\begin{figure}[]
  \centering
  \subfigure[]
  {\includegraphics[scale=0.1295]{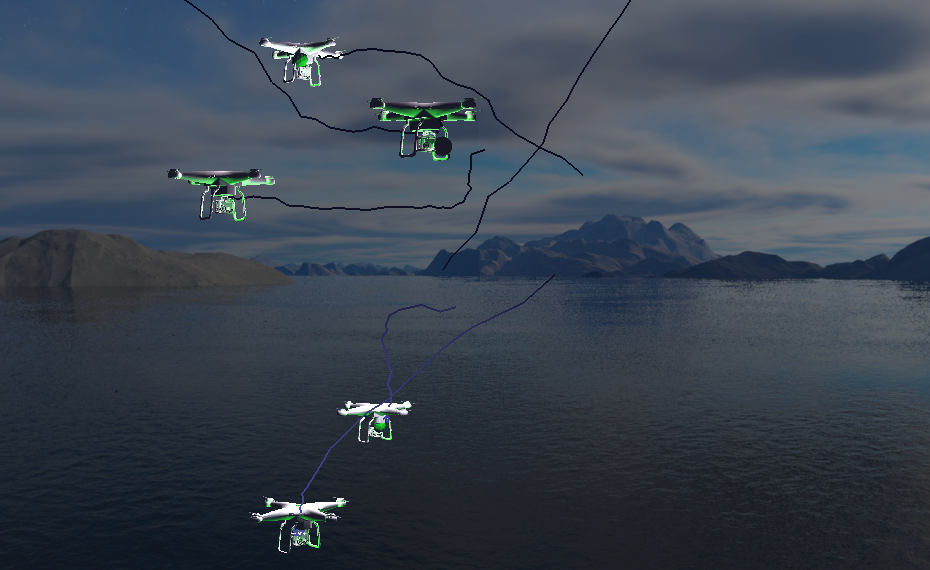}}\quad
  \subfigure[]
  {\includegraphics[scale=0.1295]{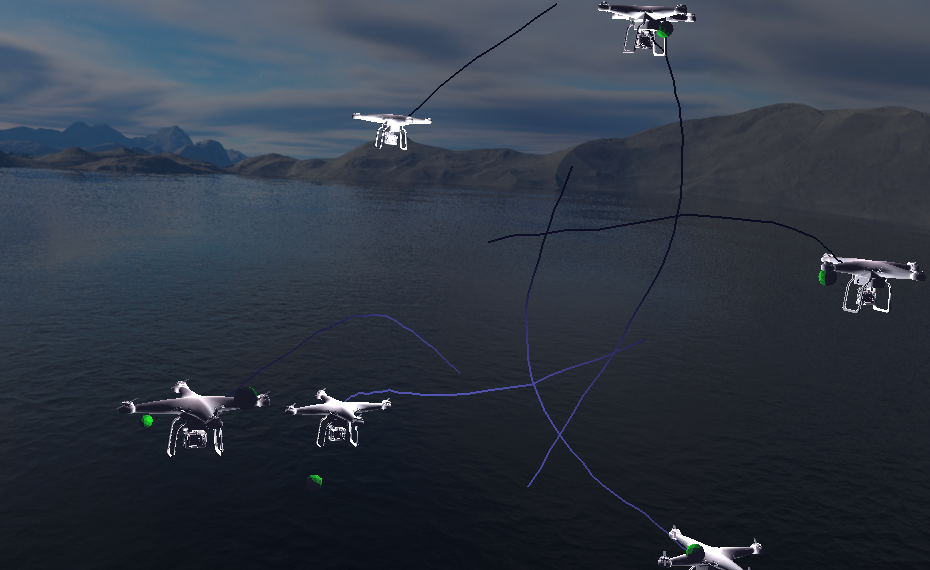}}\quad
    \subfigure[]
  {\includegraphics[scale=0.173]{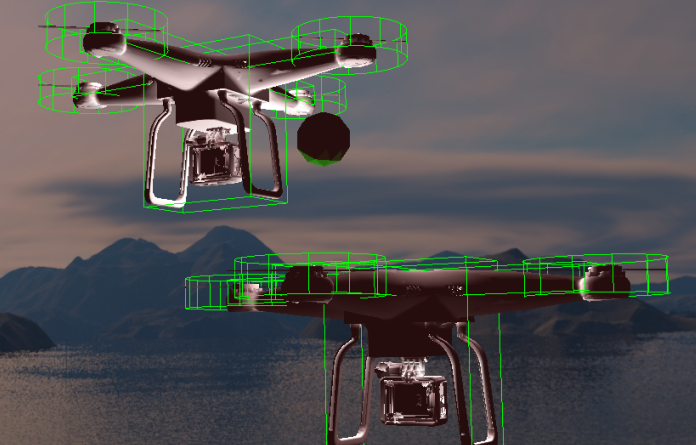}}\quad
  \subfigure[] 
  {\includegraphics[scale=0.173]{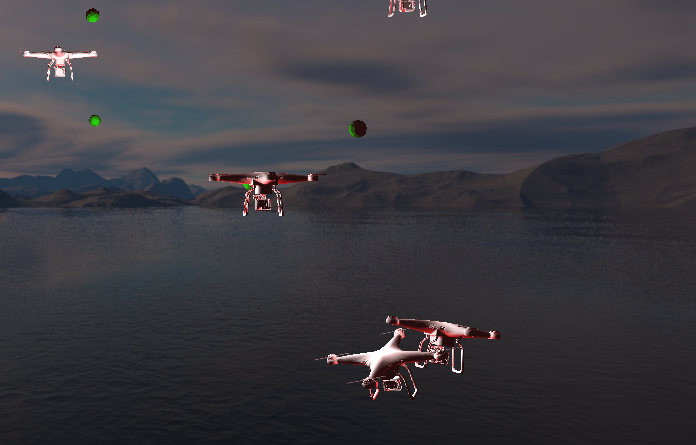}}\quad
  \caption{Coordination of multiple UAVs using Multi-Critic Policy Gradient Optimization (MCPO) to train a single actor policy network achieving collision avoidance: (a) and (b) Target navigation and rotor balancing for a number of dynamically initialized and terminated drones. (c) Simplified Collision model. (d) Collision of drones when using a single-critic architecture (Best viewed in color).}
\end{figure}

\begin{figure}[!ht]
  \centering
  {\includegraphics[scale=0.45]{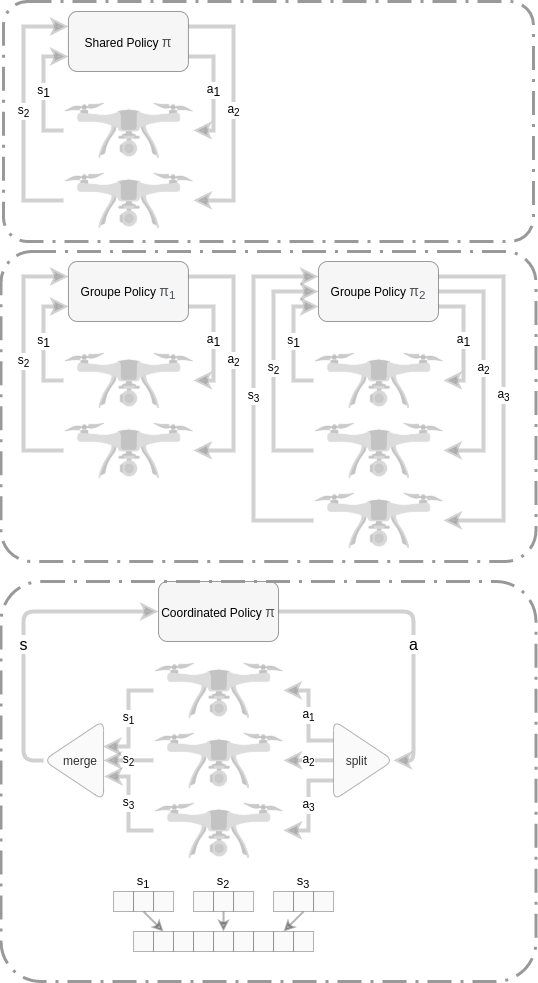}}
  \caption{Illustration of common architectures for multi-agent reinforcement learning. The top image shows a shared policy where all the agents share the same actor-policy network. For an equal state, they will behave identically. The middle image represents another architecture that applies different policies to respective groups of agents. In both cases, real coordination of agents' actions is usually achieved using various forms of communication as part of the action space. In the third and preferred architecture, coordination is an intrinsic characteristic of the architecture where a single actor policy can extract an action vector that is split into multiple sub-actions that are fed back to the agents. Their respective input state is merged from the individual states of all the agents. One of the challenges in such an architecture is to dynamically add or remove agents in an episode. In such a case, a multi-value critic-network may cause the terminated agents previously learned value estimation to deteriorate.}
\label{fig:PolicyAgentStructures}
\end{figure}

For classical multi-agent reinforcement learning, multiple approaches have been proposed to link between policies and agents \cite{ScalarizedMORL}. In a basic form, one can assign the same policy network to all agents, which would behave equally in an equivalent state. When implementing group policies, we assign a group of agents to a policy network and a different group to another policy. 
Both approaches will not achieve high-level cooperation, which for example could be partially overcome by implementing a communication component as part of the policy \cite{CommMarl, CommMarl2}. In our approach, we implement a single actor policy network that extracts multiple action vectors and feeds them to individual agents (see Figure \ref{fig:PolicyAgentStructures}). In such a way, coordination is learned as part of a single action network being aware of all the sub-states. The implementation of such a network can be accomplished by merging states and splitting actions or multi-head and multi-tail networks. In practice, common policy gradient optimization algorithms fail to train weights for such architectures, and we, therefore, need to analyze the composition and workflow of these architectures and the mechanisms involved when dealing with multiple reward signals. 

Typical actor-critic structures used in multiple policy gradient approaches \cite{ActorCriticMethods} are based on an actor policy network that receives a state and then extracts an action. The environment's reward signal is then fed into a critic value estimating network. The advantage estimation of two consecutive states is then used to train the actor policy. In the case of multi-agent reinforcement learning with the attempt to train a single actor policy network, we witness that for the typical environment setup with a single reward extraction, common approaches add up rewards to a single scalar. In this way, we lose the contribution of the individual rewards in gradient descent. We will discuss the consequence of the problem later on. Additionally, the problem of multiple reward signals is further discussed in the supplementary materials.

When approaching a reinforcement learning based drone control based on neural networks, the action space is usually continuous rather than discrete, where the rotor control allows continuous thrust. By a difference in the thrust of individual rotors, a drone controls its movement. For that reason, the choice of a policy gradient approach such as proximate policy optimization \cite{PPO} with competitive performance for most continuous approaches, is advisable. 

Reinforcement learning optimization algorithms classically are divided into value- and policy gradient-based approaches. Most modern actor-critic architectures are a hybrid form of these approaches, where an actor-network is trained based on policy gradients and a critic network is used to estimate a state value. The advantage is then extracted comparing the values of current and previous states and the actor-network is directly trained on a pessimistic advantage gradient. For the case of multiple agents, the value estimation of the merged state does not enable policy gradient-based training of an actor-network. When training the described architecture where one actor policy controls multiple agents, we find that for two or three agents training is slow, but for an increasing number of agents, the policy fails to improve and even deteriorates system performance. Moreover, due to the inability of a single critic network to trace errors originated in distinct reward signals, the actor policy optimization favours lucrative updates and neglects adverse agents. We will later use the symmetry of reward signal development as a metric to evaluate the quality of system training. 

It becomes obvious that reward signals have to be fed directly to respective networks, and that we must enable an architecture to directly back-propagate errors to their origin. The first step is to use multiple critics, that can be instantiated dynamically. For each agent, a critic is fed with the agents' reward signals and thus learns to estimate the state parameters more precisely than what a common network could do. Then, we have to enable the training of an actor policy network creating an advantage function that takes into account all the numeric estimations. In this way, weight updates can be performed to achieve an overall advantage for multiple critics. 

\subsection{Our new contributions include:} 

\begin{itemize}
  \item Achieving a high level of coordination using only a single actor policy network instead of a single policy for a group of agents.
  \item Scaling policy gradient based multi-agent reinforcement learning for a higher number of agents.
  \item Proving the system performance on a physics-based reinforcement learning environment for multiple agents with various coordination tasks.
  \item Achieving a high symmetry in reward signal development that indicates effective usage of actor-network neurons and avoidance of duplication for similar tasks. 
\end{itemize}

\section{Recent Work}
In this section, we will review the previous work performed in the area of continuous UAV control as well as multi-agent control with regards to policy gradient algorithms. 

An attempt to apply the major policy gradient algorithms on continuous UAV controls has been made in \cite{RLUAVAttitue}, which uses an open-source high-fidelity simulation environment to train a flight controller for the attitude control of a quad-copter through reinforcement learning. In consequence, accurate attitude control may be achieved similarly to our balancing task, and the implementation of proximal policy optimization is shown as the best performing for this task in most of the metrics.
The approach of \cite{RLAutonomousUAV} proposes a function approximation based reinforcement learning algorithm to deal with a large number of state representations and to obtain faster convergence. Both simulated and real implementations are provided. 
As \cite{ControllerDesign} suggests, a nonlinear autopilot for quad-rotor UAVs based on feedback linearization. The UAV control for tracking a moving target is discussed in \cite{LearningUAVControl} where a hierarchical approach combines model-free policy gradient optimization with a conventional proportional integral-derivative (PID). The approach cannot be considered as pure reinforcement learning as it undertakes supervised training of a convolutional neural network from raw images, and then applies reinforcement learning from games of self-play. To address the problem of UAV navigation in an unknown environment, the approach of \cite{ObstacleAvoidance} focuses on the combination of imitation learning and reinforcement learning based on twin delayed Deep Deterministic Policy Gradient (DDPG).

A survey of recent multi-agent approaches in reinforcement learning is provided in \cite{MultiAgentRL} with classical approaches discussed in \cite{MultiAgentOverview}. The review paper \cite{DeepMARL} provides insights into various multi-agents algorithms and addresses an important aspect of deep reinforcement learning related to the situations that require multiple agents to communicate and cooperate to solve complex tasks. In \cite{CooperativeUAVS}, the UAVs execute sensing tasks through cooperative sensing and transmission and use a compound-action actor-critic network for trajectory optimization. Synchronization, connectivity management, path planning, and energy simulation for a fleet of UAVs is discussed in \cite{UAVFleet}, and thus a control and monitor platform enables cooperative path planning by introducing swarm functions. Another interesting approach for multiple drone coordination is reported in \cite{CooperationDistributed} where full coverage of an unknown field of interest is provided by UAVs while minimizing the overlapping sections in the field of view. Their system mainly addresses the complex dynamics of the joint-actions and the challenge of a large dimensional state space. In \cite{MultiUAVNetworks}, the quality of experience-driven deployment and dynamic movement of multiple unmanned aerial vehicles are investigated and a Q-Learning based approach is proposed.

\section{Preliminary Work}

\subsection{Actor-Critic Policy Gradient Optimization}
The two main branches of reinforcement learning algorithms are value- and policy gradient-based algorithms. Generally, Policy gradient approaches are used with a continuous action space. Policy gradient algorithms have brought several advantages to deep value-based approaches such as the smooth update of policy at each step, being more effective in high dimensional action spaces with the ability to learn stochastic policies \cite{DoomPolicyGradientsQ}. The ability to learn stochastic policies is essential to learning motion in a partially observable Markov decision process and there is no need to manually implement an exploration/exploitation trade-off for the stochastic policy creates a probability for each action, making it less likely to choose the same action for a state. It also handles perceptual aliasing, when having two states that seem to be identical. 

For classical value-based approaches, the optimization target has been defined, where the loss refers to a squared error $\mathbb{E}[(G-V(s))^2]$. In Policy gradient approaches, we instead estimate the gradient of the loss and hence the  policy gradient. Then, the objective is the integral of the gradient. The formal derivation and proof of the policy gradient theorem can be found in \cite{policyGradientsLilianWeng} whereas the gradient of the target function for parameters is given as: 
\begin{equation}
\nabla_\theta J(\theta) 
= \nabla_\theta \sum_{s \in \mathcal{S}} d^\pi(s) \sum_{a \in \mathcal{A}} Q^\pi(s, a) \pi_\theta(a \vert s) 
\end{equation}
by excluding the derivative of the state distribution $d^\pi(s)$, we enable simplified computation as: 
\begin{equation}
\propto \sum_{s \in \mathcal{S}} d^\pi(s) \sum_{a \in \mathcal{A}} Q^\pi(s, a) \nabla_\theta \pi_\theta(a \vert s)
\end{equation}
In simple terms, the estimation of Advantages times the gradient of the logarithm of the policy distribution: 
\begin{equation}
\nabla_\theta J(\theta) = \mathbb{E}[A(s,a)\nabla_{\theta} \log \pi(a|s)]
\end{equation}
which leads to the estimation as follows:
\begin{equation}
\nabla_\theta J(\theta) \approx \frac{1}{N} \sum_{i = 1}^N [A(s,a)\nabla_{\theta} \log \pi(a|s)]
\end{equation}

\subsection{Modern policy optimization}
Trust Region policy optimization \cite{TRPO} tends to constrain the parameter update of the policy using 
Kullback–Leibler (KL) divergence \cite{KL, KL2}. It aims to maximize the objective function \cite{policyGradientsLilianWeng}:
\begin{equation}
J(\theta) = \mathbb{E}_{s \sim \rho^{\pi_{\theta_\text{old}}}, a \sim \pi_{\theta_\text{old}}} \big[ \frac{\pi_\theta(a \vert s)}{\pi_{\theta_\text{old}}(a \vert s)} \hat{A}_{\theta_\text{old}}(s, a) \big]
\end{equation}
where the distance between the old and new policy represented by the KL-distance is constrained to be within $\delta$:
\begin{equation}
\mathbb{E}_{s \sim \rho^{\pi_{\theta_\text{old}}}} [D_\text{KL}(\pi_{\theta_\text{old}}(.\vert s) \| \pi_\theta(.\vert s)] \leq \delta
\end{equation}
As another advantage in policy gradient optimization, ACER \cite{ACER} is an actor-critic approach using experience replay. Based on the asynchronous advantage actor-critic approach \cite{A3C}, ACER acts as an off-policy. It uses retrace Q-Value estimation, applies efficient TRPO, and truncates importance weights with bias correction \cite{policyGradientsLilianWeng}.
The Soft Actor-Critic Algorithm (SAC) \cite{SAC} encourages exploration by incorporating an entropy measure of the policy into the reward. Similar to proximal policy optimization \cite{PPO}, Soft Actor-Critic uses a policy network and a value estimation network with entropy maximization to achieve stability and exploration  \cite{policyGradientsLilianWeng} :
\begin{equation}
J(\theta) = \sum_{t=1}^T \mathbb{E}_{(s_t, a_t) \sim \rho_{\pi_\theta}} [r(s_t, a_t) + \alpha \mathcal{H}(\pi_\theta(.\vert s_t))]
\end{equation}

The policy update then is formed as:
\begin{equation}
\Delta \theta = \alpha \nabla_{\theta}(\log \pi_{\theta}(s,a)) \hat{q}_{\omega}(s,a)
\end{equation}

Using the advantage, the value function is formed: 
\begin{equation}
\Delta \omega = \beta (R(s,a) + \gamma \hat{q}_{\omega}(s_{t+1},a_{t+1}) - \hat{q}_{\omega}(s_t,a_t))  \nabla_{\omega} \hat{q}_{\omega}(s,a)
\end{equation}
Actor and critic policies have different learning rates, i.e. $\alpha$ and $\beta$, while the critic network usually updates faster. The trajectory probabilities multiplied with the specific trajectory reward are summed up to:
\begin{equation}
J(\theta) = \sum_{s \in \mathcal{S}} P^\pi(s) V^\pi(s) 
= \sum_{s \in \mathcal{S}} P^\pi(s) \sum_{a \in \mathcal{A}} \pi_\theta(a \vert s) Q^\pi(s, a)
\end{equation}
where $P^{\pi}$ is the stationary distribution of the Markov chain for $\pi$.

Using the probability ratio as that used in proximate policy optimization (PPO) \cite{PPO}, we restrict the ratio $r_t(\theta)$ to the form as:
\begin{equation}
r(\theta) = \frac{\pi_\theta(a \vert s)}{\pi_{\theta_\text{old}}(a \vert s)}
\end{equation}
So, the objective function becomes: 
\begin{equation}
L^\text{CPI} (\theta)= \mathbb{E} [ \frac{\pi_\theta(a \vert s)}{\pi_{\theta_\text{old}}(a \vert s)} \hat{A}_t) =  \mathbb{E} [ r(\theta) \hat{A}_t) ]
\end{equation}
where $L^\text{CPI}$ is the conservative policy iteration as described in \cite{CLI}.

\begin{equation}
L^\text{CLIP} (\theta) = \mathbb{E}_t [ \min( r_t(\theta) \hat{A}_t, \text{clip}(r_t(\theta), 1 - \epsilon, 1 + \epsilon) \hat{A}_t)]
\end{equation}
The new term is of: 
\begin{equation}\text{clip}(r(\theta), 1 - \epsilon, 1 + \epsilon)\end{equation}
which clips the value of $r_t(\theta)$ in the interval $[1-\epsilon, 1 + \epsilon]$ depending on two cases $\hat{A}_t > 0$ and $\hat{A}_t \leq 0$, where the first case $\hat{A}_t > 0$ and the value of $r_t(\theta)$ increases leading to a higher probability for this action. The clipping for $r_t(\theta)$ by $1 + \epsilon$ that limits the update. In the case of $\hat{A}_t \leq 0$, no advantage is recognized and the action should not be adopted. By clipping the value of $r_t(\theta)$ by $1-\epsilon$, the decrease is not drastic. The loss function is extended by a term for exploration as  described in A3C \cite{A3C}. The final Loss is thereby constructed with squared error loss between the actual and target value functions and the entropy element.
\begin{equation}
L^\text{CLIP'} (\theta) = \mathbb{E} [ L^\text{CLIP} (\theta) - c_1 (V_\theta(s) - V_\text{target})^2 + c_2 H(s, \pi_\theta(.)) ]
\end{equation} 
We here briefly review the approach of Deep Deterministic Policy Gradient Optimization (DDPG) \cite{DDPG}, an actor-critic and model-free algorithm that is based on DPG and DQN and enables continuous controls on a deterministic rather than stochastic basis. On this basis, Multi-Agent Actor-Critic for Mixed Cooperative-Competitive Environments \cite{MADDPG} proposes actor-critic methods that consider action policies of other agents, and the groups of agents learn to discover
physical and informational coordination strategies using the actor update:
$$
\nabla_{\theta_i} J(\theta_i) =
$$
\begin{equation}
\mathbb{E}_{\vec{o}, a \sim \mathcal{D}} [\nabla_{a_i} Q^{\vec{\mu}}_i (\vec{o}, a_1, \dots, a_N) \nabla_{\theta_i} \mu_{\theta_i}(o_i) \rvert_{a_i=\mu_{\theta_i}(o_i)} ]
\end{equation}

\subsection{Recent policy optimization innovations}
In recent developments of policy optimization, one has to mention constrained policy optimization \cite{ConstrainedPolicyOptimization} where both, reward function and constraints are specified. Thereby, safety constraints are not specified as part of the reward signal as typically done. This approach guarantees to constraint satisfaction at each iteration. Another relevant development in Policy Gradient approaches is Model-Ensemble Trust-Region Policy Optimization \cite{ModelEnsembleTRPO} that proposes to use an ensemble of models to maintain the model uncertainty and regularize the learning process. It is shown that the use of likelihood ratio derivatives yields much more stable learning than back-propagation over time. In recent days, another approach of  Model-Based Meta-Policy-Optimization  \cite{Model-Based} is constructed to be more robust to model imperfections than previous model-based approaches, using an ensemble of the learned dynamic models.

\section{Proposed Method}
Using an established policy gradient approach, we enhance the advantage function by value estimations of multiple critics being able to back propagate the errors to individual critics and therefore effectively train an actor policy. First, we derive the choice of gradient optimization used for this approach. 
\subsection{Choice of policy gradient optimization}
The choice of a policy gradient approach is based on its ability to handle high-dimensional continuous action and state spaces, their good convergence and the ability to learn a stochastic policy. The objective of improving a differentiable actor policy $\pi_{\theta}(s,a)$ depending on the actor-networks' parameters $\theta$. The quality $J(\theta)$ of policy $\pi_{\theta}(s,a)$ is based on the value estimation. For the initial state, we have $J(\theta) = V^{\pi_{\theta}}(s)$. Given the state probability $d^{\pi_{\theta}}$, the summed reward from a starting state  can be written as: 
\begin{equation}
J(\theta) = \sum_s d^{\pi_{\theta}}(s) V^{\pi_{\theta}}(s)
\end{equation}
With the value factor derived from policy $\pi$ and respective state-action rewards $R(s,a)$, we obtain: 
\begin{equation}
J(\theta) = \sum_s d^{\pi_{\theta}}(s) \sum_a \pi_{\theta}(s,a)R(s,a)
\end{equation}
Which is defined for a single value estimation with a single reward source. The localization $\theta$ is determined as the one that maximizes $J(\theta)$ using gradient ascending. As has been known, the parameter update can be written as $\Delta\theta = \alpha \nabla_{\theta}J(\theta)$ with learning rate $\alpha$.  As previously indicated, the policy gradient theorem now states that:
\begin{equation}
\nabla_{\theta}\pi_{\theta}(s,a) = \pi_{\theta}(s,a)\nabla_{\theta} \log \pi_{\theta}(s,a)
\end{equation}
Thus we have: 
\begin{equation}
\nabla_{\theta} J(\theta) = \mathbb{E} [\pi_{\theta}(s,a)\nabla_{\theta} \log \pi_{\theta}(s,a) Q^{\pi_{\theta}}(s,a)]
\end{equation}
Where the reward is replaced by a state-value estimation. The policy gradient theorem is valid for different objective functions as well. Using a critic to estimate the state value function helps us to establish the group actor-critic algorithms. We use the common gradient estimator's form:
\begin{equation}
\mathbb{E} [\nabla_{\theta} \log \pi_{\theta}(s,a) A_t]
\end{equation}
where $A_t$ is introduced as the estimator of the advantage function. This results in the loss functions of the form: 
\begin{equation}
L(\theta) = \mathbb{E} [\log \pi_{\theta}(s,a) A_t]
\end{equation}
We use a surrogate loss based on the ratio $r(\theta)$ of the current $\pi_\theta(a \vert s)$ and the previous $\pi_{\theta_\text{old}}(a \vert s)$ policy network parameters where we maximize:
\begin{equation}
\mathbb{E}[r(\theta)A(s,a)] = \mathbb{E} [\frac{\pi_\theta(a \vert s)}{\pi_{\theta_\text{old}}(a \vert s)}A_t)]
\end{equation}
The advantage is now based on the value estimation of current and previous states $s', s$ with $\gamma$ as the discount factor. For the original advantage based on the state-action value $Q(s,a)$:
\begin{equation}
A(s,a) = Q(s,a) - V(s)
\end{equation}
A good approximation is:
\begin{equation}
A(s,a) \approx r + \gamma V(s') - V(s)
\end{equation}

\begin{figure}[]
  \centering
  {\includegraphics[scale=0.41]{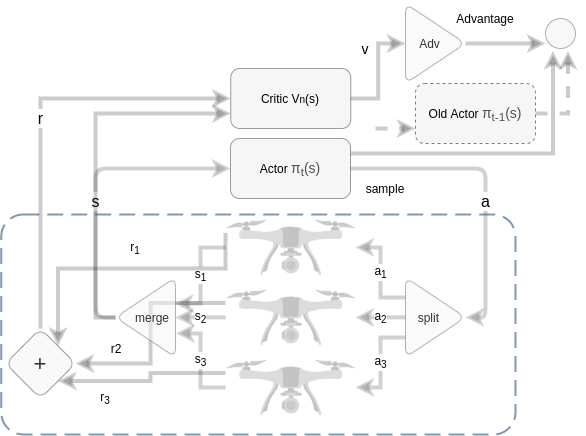}}
  \caption{A single critic architecture that is adapted for multiple agents that adds up reward signals and exports them as scalar to its critic. }
  \label{architectureSingle}
\end{figure}

\subsection{Standard Single-Critic Scalar Advantage}
Compared to our approach, a standard actor-critic approach (see Fig. \ref{architectureSingle}) with a single critic $C$ that extracts a single value estimation for each state $V(s)$:
\begin{equation}
C \to V(s)
\end{equation}
with a scalar $\dim_0(A) = 1$ advantage: 
\begin{equation}
A(r, V(s), V(s'))
\end{equation}
which is based on a black box environment that just extracts a single reward signal $r$. The policy probability ratio is  multiplied with the corresponding scalar advantage: 
\begin{equation}
r(\theta)A(s,a) = \frac{\pi_\theta(a \vert s)}{\pi_{\theta_\text{old}}(a \vert s)}(r + \gamma V(s') - V(s))
\end{equation}
The environment extracts a single reward signal and therefore errors cannot be back-propagated correctly to achieve an efficient optimization of a policy for a higher number of agents.

\subsection{Extracting multiple reward signals with multiple critics.}

In our attempt to backpropagate errors by individual agents, it becomes clear that we have to extract multiple reward signals from the environment in a way that either a reward vector is fed to a single critic $C(r^i)$, like that used in our hybrid approach to be discussed later:
\begin{equation}
r^i \to C
\end{equation}
Or, in a way that each reward signal is assigned to a  dynamically initiated critic:
\begin{equation}
r^i \to C^i : r^1 \to C^1, r^2 \to C^2, ..., r^n \to C^n
\end{equation}
Each critic then extracts a value estimation for its corresponding agent: 
\begin{equation}
C^i \to V^i(s) : C^1 \to V^1(s), C^2 \to V^2(s), ..., C^n \to V^n(s)
\end{equation}

\begin{figure}[]
  \centering
  {\includegraphics[scale=0.45]{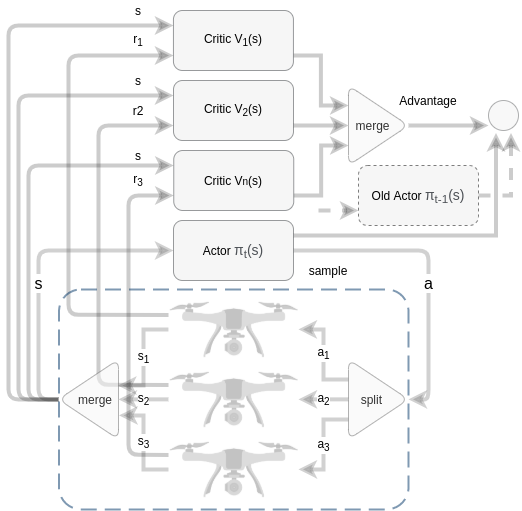}}
  \caption{The proposed Poliymorph architecture feeds individual reward signals to corresponding critics, where  their architectures may not be necessarily equal. The distinct training of value estimation networks leads to a better actor policy update. }
  \label{architectureMulti}
\end{figure}

Contrary to a single-critic architecture, we can now instantiate each critic in the runtime (see Figure \ref{architectureMulti}), Where critics do not necessarily have to be of an identical structure. Some agents may require a different design or depth to achieve more efficient value estimation. Convectional value estimation of a single critic is accomplished using a single reward signal $r_t$, where the accumulated reward is defined by the sum of all the sub-rewards,  $r_t = r^1_t + r^2_t + ...+ r^n_t$ where we feed the environment rewards to their critics directly. Each critic value estimation network $C_i$ is fed with a full state vector and a reward signal $C_i(\textbf{s}, r_i)$. Naturally, we can assume that a single critic using the accumulated reward would be effectively trained. Nevertheless, for the reasons explained earlier and confirmed by our experiments, the performance of such architecture does not scale for a large number of agents. One has to consider that the policy gradient optimization performed for the actor policy network has to backpropagate errors to changes in reward signals. This ability is lost when reward signals are simply accumulated and returned as a black box. As known, the value function $V_{\pi}(s)$ for actor policy $\pi$ is based on the expected future rewards. We have
\begin{equation}
V_{\pi}(s) =\mathbb{E}_{\pi}[R_t(r^1_t, r^2_t, ..., r^n_t) | s_t = s]
\end{equation}
which is of $n$ critics. 
\begin{equation}
V_{\pi}(s) =\mathbb{E}_{\pi}[R_t | s_t = s]
\end{equation}
To derive an optimal policy $\pi^{*}$, we formulate the optimized value function $V^{*}(s)$ as:
\begin{equation}
V^{*}(s) = max_{\pi} \{ V_{\pi} \} , \forall s \in S
\end{equation}
which will usually never be achieved for a single-critic architecture due to neglection of sub-objectives as described earlier. A distinct estimation using multiple critics extracts individual value functions for the same policy $\pi$:
$$
V^1_{\pi}(s) =\mathbb{E}_{\pi}[R_t(r^1_t) | s_t = s]
$$
$$
\vdots 
$$
$$
V^n_{\pi}(s) =\mathbb{E}_{\pi}[R_t(r^n_t) | s_t = s]
$$
The direct objective of maximizing the overall value is the same as that of the Bellman equation:
\begin{equation}
V(s) = \mathbb{E}[R_{t+1} + \gamma V(S_{t+1} | S = s_t)]
\end{equation}
With optimal policy $\pi^*$, we have the satisfactory accumulated value:
\begin{equation}
V^{*}(s) = max_{\pi} \{ V_{\pi} \} , \forall s \in S
\end{equation}
As stated, optimal value estimation $V^{i*}$ cannot be achieved by the optimizer. Instead, one should aim to optimize value estimation $V^i$ for all the sub-objectives with: 
\begin{equation}
V^i(s) = \mathbb{E}[\sum_{i = 0}^ nR^i_{t+1} + \gamma V(S_{t+1} | S = s_t)]
\end{equation}
Usually, a buffer of rewards is discounted according to:
\begin{equation}
G_t = \sum_{k=0}^n \gamma^k R_{t+k+1}
\end{equation}
with $\gamma$ as the discount factor, and a penalty of future rewards $0 < \gamma \leq 1$.

\subsection{Multi-Critic Advantage}
The optimization of critics is based on a simple mean square error loss where we estimate the individual values per state  $V^i(s_{t+1})$. The advantage of $A(s,a) = Q(s,a) - V(s)$ is approximated using: 
\begin{equation}
A(s,a) \approx r + \gamma V(s') - V(s)
\end{equation}
However, we use multiple critics for value extraction in a way that each critic extracts a distinct value estimation (see Figure \ref{architectureMulti}): 
\begin{equation}
C^1 \to V^1(s), C^2 \to V^2(s), ...
\end{equation}
These critics are fed with a full state $C^i(r^i)$ and individual rewards:
\begin{equation}
r^i \to C^i
\end{equation}
The advantage now depends on:
\begin{equation}
A(r^1, r^2, ..., V^1(s), V^2(s), ..., V^1(s'), V^2(s'), ...)
\end{equation}
where, for multiple critics $C^i$ with $i \in \{1,2,..n\}$, we obtain:
\begin{equation}
A(s,a) \approx r^i + \gamma V^i(s') - V^i(s)
\end{equation}
We consider each argument as a vector of critics.
The optimizer now requires equal dimensions of advantages and actor policy probability ratio depending on the parameter $\theta$:
\begin{equation}
r(\theta) = \frac{\pi_\theta(a \vert s)}{\pi_{\theta_\text{old}}(a \vert s)}
\end{equation}
The advantages can be stacked from the action dimension of an agent $\dim(a_0)$:
\begin{equation}
\langle A(s,a) \mid \odot \rangle_{\dim(a_0)} := \begin{bmatrix} \vdots \\ A(s,a) \\ \vdots \end{bmatrix}_{\dim(a_0)} 
\end{equation}
So the dimension of the original advantage is: 
\begin{equation}
\dim(a_0) \cdot \dim_0(A) = \dim_0(r(\theta))
\end{equation}
Then, the stacked advantage becomes the surrogate: \begin{equation}
r(\theta) \langle A(s,a) \mid \odot \rangle_{\dim(a_0)} = \frac{\pi_\theta(a \vert s)}{\pi_{\theta_\text{old}}(a \vert s)} \langle A(s,a) \mid \odot \rangle_{\dim(a_0)}
\end{equation}
In this way, the optimizer can back propagate errors to each advantage based on individual value estimations. So, the actor policy loss function becomes:

\begin{equation}
L^\text{A} (\theta)= \mathbb{E} [ \frac{\pi_\theta(a \vert s)}{\pi_{\theta_\text{old}}(a \vert s)} \langle A(s,a) \mid \odot \rangle_{\dim(a_0)} ] 
\end{equation}
\begin{equation}
=  \mathbb{E} [ r(\theta) \langle A(s,a) \mid \odot \rangle_{\dim(a_0)} ]
\end{equation}

Now, with clipping, the loss becomes:

\begin{equation}
L^\text{C} (\theta) = \mathbb{E}_t [ \min( r_t(\theta) \langle A(s,a) \mid \odot \rangle_{\dim(a_0)}), 
\end{equation}
\begin{equation}
\text{clip}(r_t(\theta), 1 - \epsilon, 1 + \epsilon) \langle A(s,a) \mid \odot \rangle_{\dim(a_0)}))]
\end{equation}
The new term 
\begin{equation}\text{clip}(r(\theta), 1 - \epsilon, 1 + \epsilon)\end{equation}
clips the value of $r_t(\theta)$ in the interval $[1-\epsilon, 1 + \epsilon]$, depending on two cases $\hat{A}_t > 0$ and $\hat{A}_t \leq 0$. Where the first case $\hat{A}_t > 0$, the value of $r_t(\theta)$ increases, leading to a higher probability for this action. The clipping for $r_t(\theta)$ by $1 + \epsilon$ limits the update. In the case of $\hat{A}_t \leq 0$, no advantage is recognized and the action should not be adopted. By clipping the value of $r_t(\theta)$ by $1-\epsilon$, the decrease is not drastic. 

\begin{algorithm}
\SetAlgoLined
\SetKwInOut{Input}{input}\SetKwInOut{Output}{output}
\Input{Initialization of $\theta_0$ }
\BlankLine
\While{$episode \leq episodes$}{
\While{$st \leq epLengt$}{
Choose a stochastic action based on a multivariate normal distribution of actor policy network $\pi_{\theta}(s) \to \sigma, \mu$
Perform $s',r = step(a)$ and obtain estimated advantage from critics: 
$A(s,a) \approx r + \gamma V(s') - V(s)$ \\
}
\If{$bufferSize \geq batchSize$}{
\For{$critic$ $\textbf{in}$ $critics$}{
State value estimation using critic loss
$L_v = \mathbb{E}[(G-V(s;\theta_v))^2] $
where G is the episodes discounted reward, and V(s) the critics value estimation.
}
Update actor policy:
$\Delta \theta = \alpha \nabla_{\theta}(\log \pi_{\theta}(s,a)) \hat{q}_{\omega}(s,a)$ \\
with the surrogate:
$r(\theta) \langle A(s,a) \mid \odot \rangle_{dim(a_0)} $ \\
pessimistic loss becomes: \\
$L^\text{C} (\theta) = \mathbb{E}_t [ \min( r_t(\theta) \langle A(s,a) \mid \odot \rangle_{\dim(a_0)})$
}
\caption{Polymorph Policy Gradient Optimization.}
}
\end{algorithm}

\subsection{Hybrid-Critic Advantage}
On the other hand, we define a hybrid approach that is slightly different from Polymorph policy optimization using a single critic $C^h$ that will extract multiple values $v_1, v_2, ..., v_n$ from the final dense layer.
\begin{equation}
C^h \to v_1, v_2, v_3, ...
\end{equation}
All rewards $r^i$ are fed as a vector into $C^h$:
\begin{equation}
r^i \to C^h
\end{equation}
where mean square errors of the reward vector and the predicted value vector are used to optimize the critic. Now, the advantage depends on those values and the reward vector $r^i$,
\begin{equation}
A(r^1, r^2, .... v_1, v_2, v_3,...,... v_1', v_2', v_3',...) 
\end{equation}
Then, the simplified advantage is obtained by 
\begin{equation}
A(s,a) = (r^i + \gamma v_1 - v_1')
\end{equation}
where again $r$ is the reward vector depending on the number of the agents, and $V(s')$ is the extracted vector of the values from the last dense layer of the single critic $C^h$: $v_1, v_2, ..., v_n$.

Here, a complete reward vector is fed into $C^h$, different from Poliymorph optimization where each reward is just fed into its corresponding critic. Here again, we repeat the advantage in order to obtain the same dimension as the policy probability ratio because: 
\begin{equation}
\dim(a_0) \cdot \dim_0(A) = \dim_0(r(\theta))
\end{equation}
Therefore,  the final advantage becomes:
\begin{equation}
\langle A(s,a) \mid \odot \rangle_{\dim(a_0)}
\end{equation}
We then follow the same procedure as that for Poliymorph policy gradient optimization using clipping.

\subsection{High dimensional state merging versus multi-head and multi-tail networks}
We assume that a state $s$ contains the concatenated state information of all the sub-state. An architecture, that aims to achieve an optimal degree of coordination with a single actor policy network, will have to be fed with all the states by simply concatenating the agent states $s_1, s_2, ...., s_n$ to a single long state vector $\textbf{s} = \{s_1|s_2|...|s_n\}$. The resulting high dimensionality shall be another argument for the choice of policy gradient optimization. 


\begin{figure}[]
  \centering
  {\includegraphics[scale=0.45]{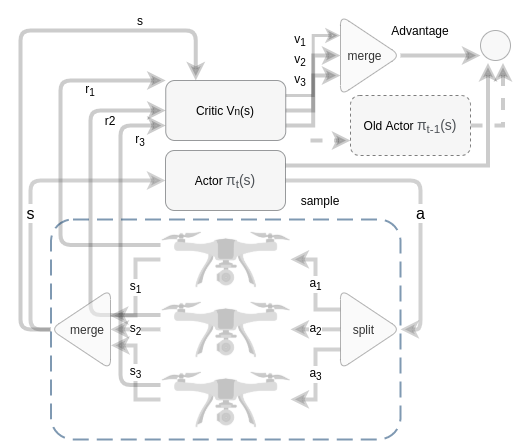}}
  \caption{Hybrid architecture with a single critic extracting multiple values. }
  \label{architectureHybrid}
\end{figure}

In practice, when we need to instantiate and terminate agents within an episode, other feeding methods may be considered, which will not deteriorate the actor policy network's performance by changing the established weights. Respectively, the actor policy network outputs a single long action vector that can be split and fed to multiple agents as shown $\textbf{a} = \{a_1|a_2|...|a_n\}$. Both can also be accomplished using a functional API to define multiple heads and tails, and for certain applications, the use of deep learning networks that allow dynamic changes of the network structure such as chainers may be considered.



\section{Experimental Work}

The objective of our experiments is to evaluate the performance of Poliymorph Policy Gradient Optimization compared to Single-Critic and Hybrid architectures. We, therefore, provide several customized multi-agent reinforcement learning environments based on multiple UAVs that return a reward vector with the respective rewards for each agent instead of a scalar reward as usually deployed in most reinforcement learning environments. We choose to utilize proximal policy optimization rather than deep deterministic policy gradients due to the continuous action space of the environments.

\begin{figure}[]
  \centering
  \subfigure[ ]{\includegraphics[scale=0.13]{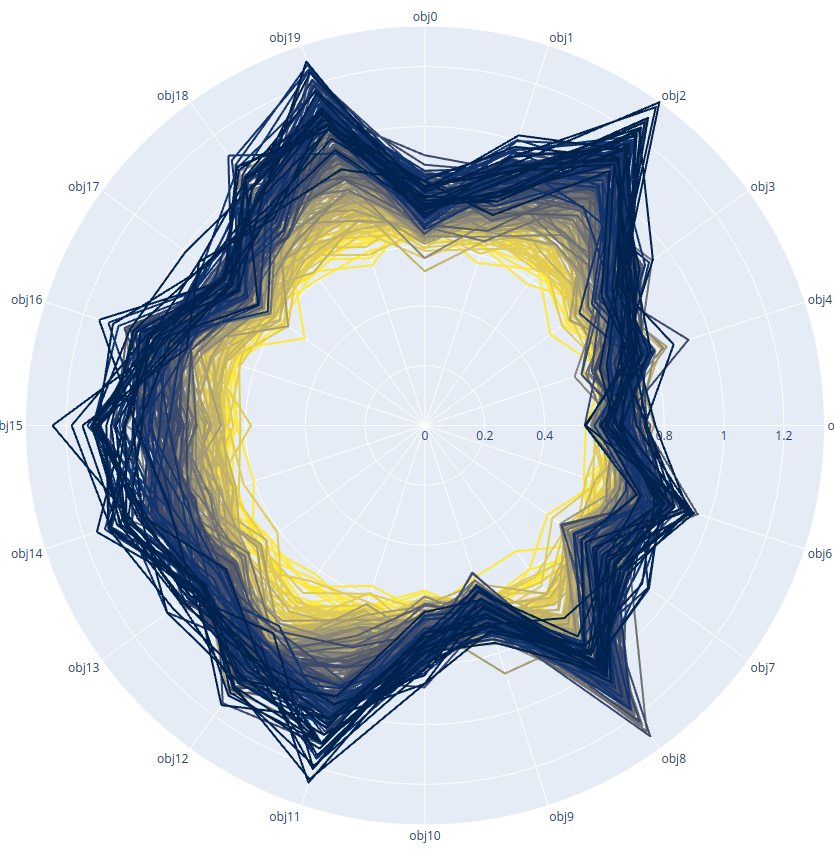}}\quad
  \subfigure[
  ]{\includegraphics[scale=0.13]{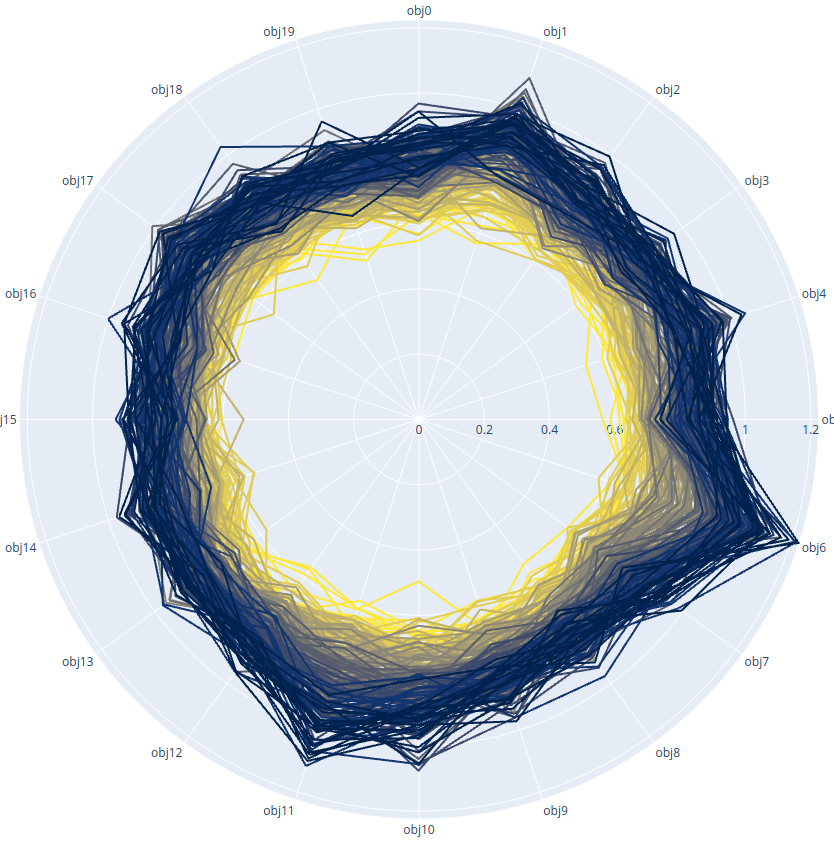}}\quad
  \caption{
  The similarity of reward signal development can be illustrated using a radar plot that represents sub-reward stimulation per episode for $n$ sub-objectives, where the inner points (yellow) stand for the mean of early training episode rewards, and blue represents later training episodes. The symmetry of the graph in this form of visualization indicates a high rate of similarity in optimization for different sub-objectives. For this example, using standard single-critic proximal policy optimization, the grave asymmetry indicates different speeds of the optimization of the sub-objectives without optimally using the actor-policy network. Similar paces of reward signal stimulation can be illustrated by a symmetric multi-objective radar plot. (a) Single-Critic using classical Proximal Policy Optimization (PPO) that estimates total values. (b) Poliymorph Policy Gradient Optimization (PMPGO) balancing value estimations of multiple critics via a recurrent parent network. The above sample shows a non-conflicting case for a number of 20 sub-objectives. Our algorithm cannot entirely eradicate pace asymmetry, given that the balancing of the recurrent parent network is performed offline but the overall advantage for the use of the actor policy network is evident (Best viewed in color). }
\end{figure}
\subsection{Physics based reinforcement learning environment}
The base quad-copter environment is a Physics-based Reinforcement Learning Environment training a quad-copter to accomplish various tasks. The python environment was created using panda3d and Bullet, a modern and open-source physics engine. The base environment is highly customizable and many tasks can be implemented and then form individual environments. These environments can be configured not to render during training and to render for the visualization purpose when the networks are trained. The GitHub code repositories for this work can be found at \cite{Multi-Quadcopter-env} and \cite{Quadcopter-env}.

A collision model can detect collisions of UAVs. Two models for motion control are implemented, one based on a 4-dimensional action space simulating a direct motor control and a simplified 3-dimensional action space control model that transmits a direction vector, assuming that firmware is able to convert this direction vector to motor differentials that implement the desired motion. All the environments are based on a standard gravity model, and additionally, wind can be simulated in various forms. 

\begin{figure}[]
  \centering
  \subfigure[n=2 agents
  ]{\includegraphics[scale=0.25]{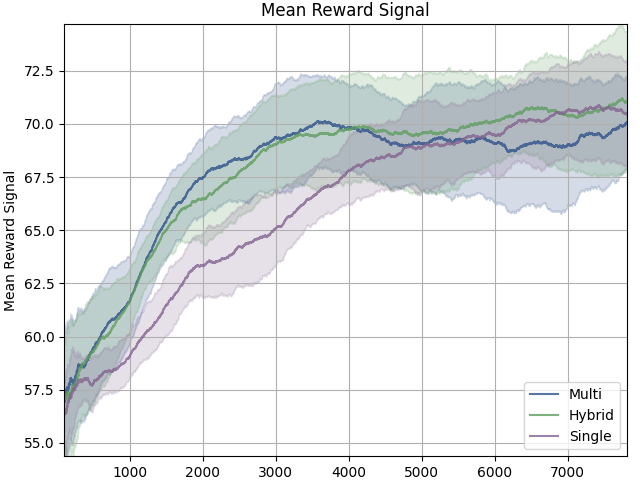}}\quad
  \subfigure[n=3 agents
  ]{\includegraphics[scale=0.25]{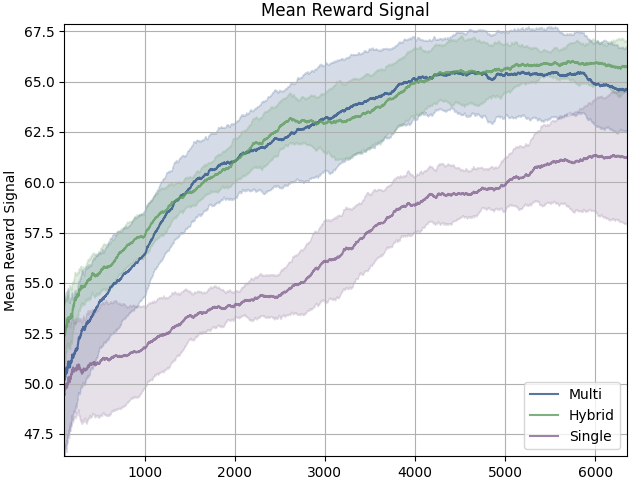}}\quad
  \subfigure[n=4 agents
  ]{\includegraphics[scale=0.25]{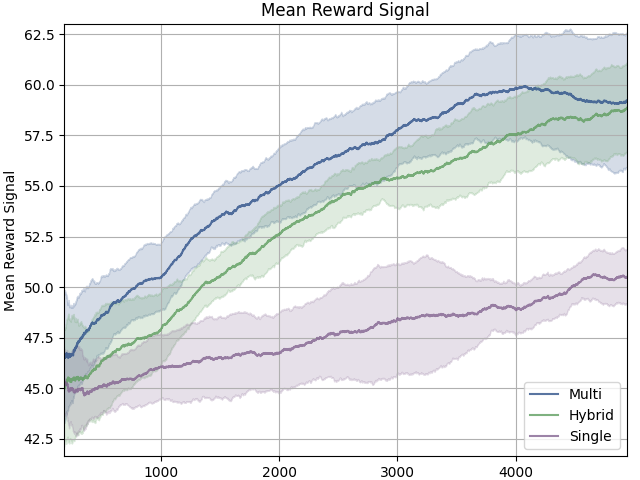}}\quad
  \subfigure[n=5 agents
  ]{\includegraphics[scale=0.25]{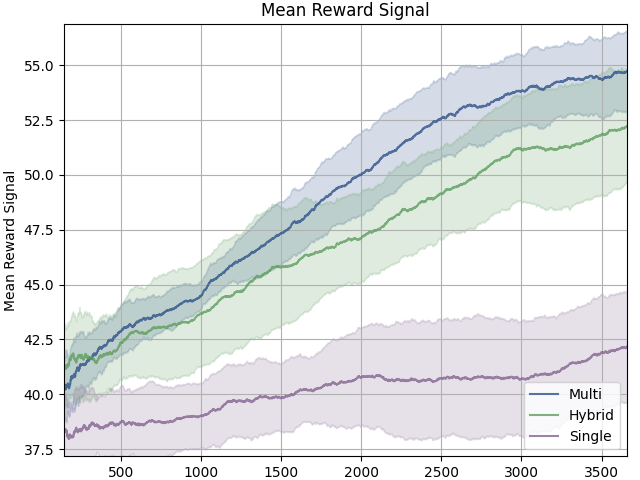}}\quad
  \subfigure[n=6 agents
  ]{\includegraphics[scale=0.25]{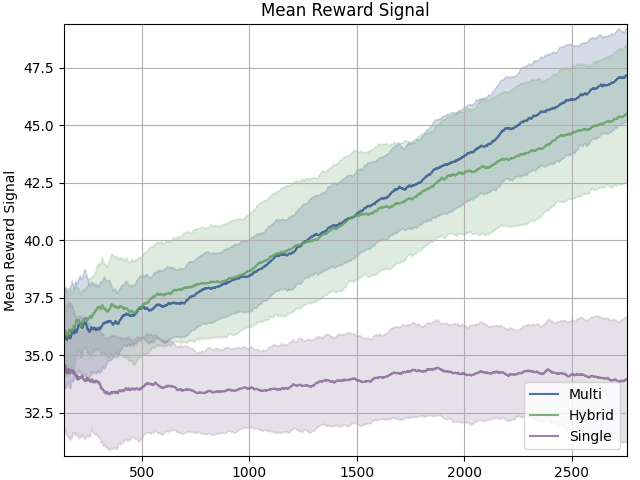}}\quad
  \subfigure[n=7 agents
  ]{\includegraphics[scale=0.25]{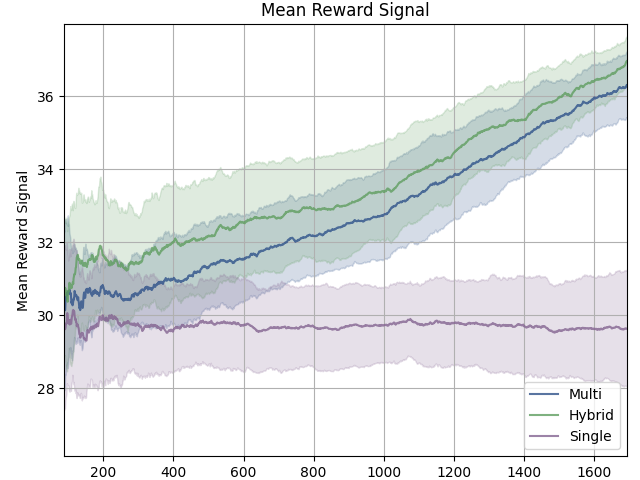}}\quad
  \caption{Smoothed mean reward signals per training episode. Each graph is conscripted of 8 training sessions per case and then averaged and plotted with its standard deviation. With an increasing number of agents, the classical single-critic architectures fails to effectively train a stable actor policy. Generally, the average of Multi-Critic reward signals exceeds that of a hybrid architecture with multi-value single-critic (Best viewed in color). }
\end{figure}

\begin{figure}[]
  \centering
  \subfigure[n=2 agents
  ]{\includegraphics[scale=0.25]{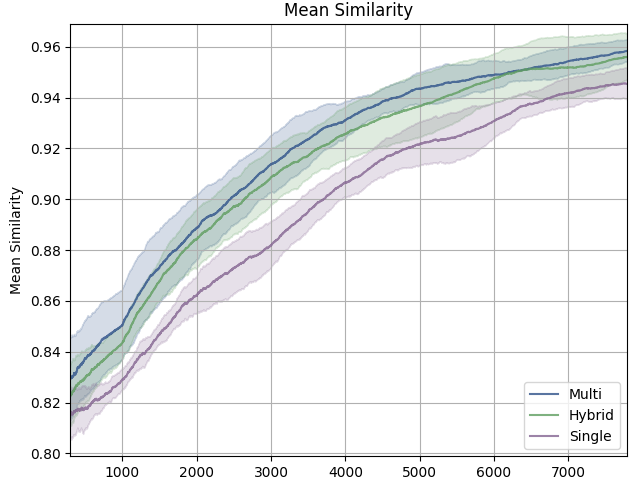}}\quad
  \subfigure[n=3 agents
  ]{\includegraphics[scale=0.25]{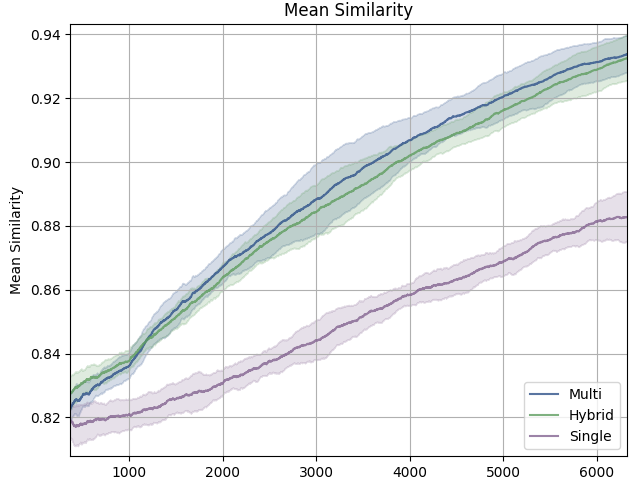}}\quad
  \subfigure[n=4 agents
  ]{\includegraphics[scale=0.25]{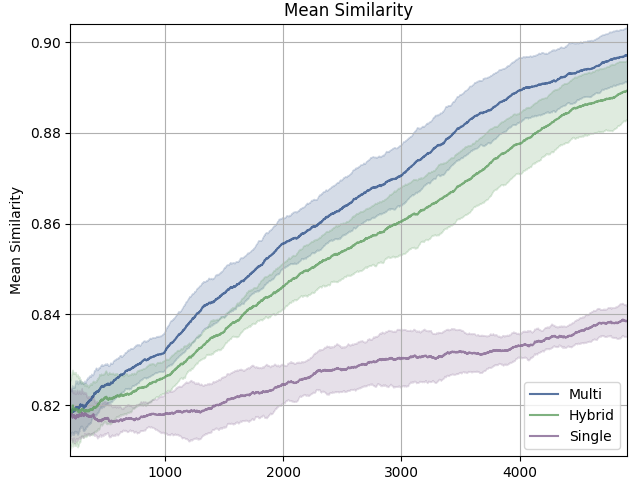}}\quad
  \subfigure[n=5 agents
  ]{\includegraphics[scale=0.25]{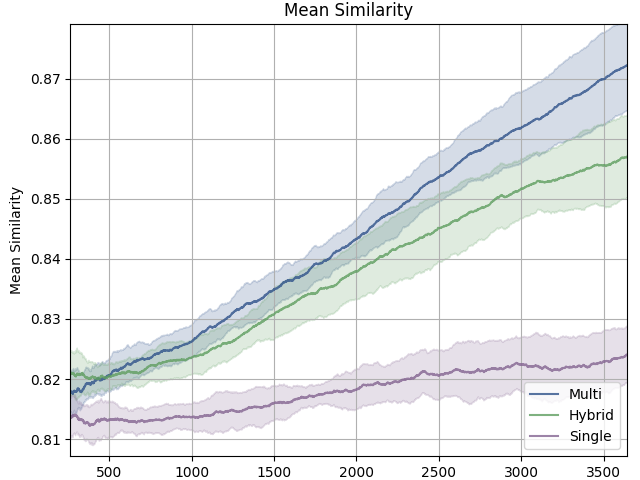}}\quad
  \subfigure[n=6 agents
  ]{\includegraphics[scale=0.25]{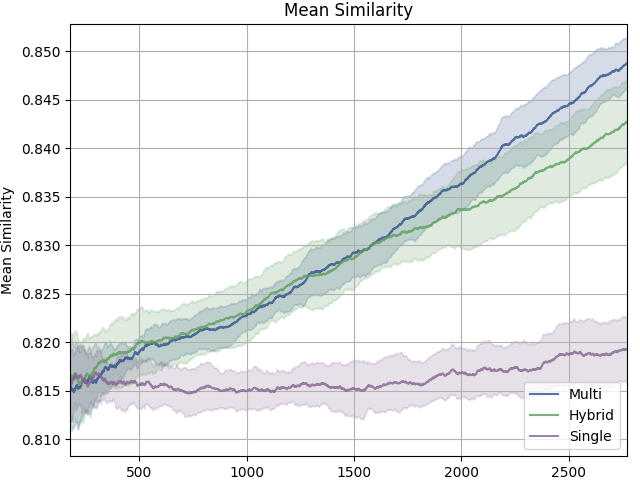}}\quad
  \subfigure[n=7 agents
  ]{\includegraphics[scale=0.25]{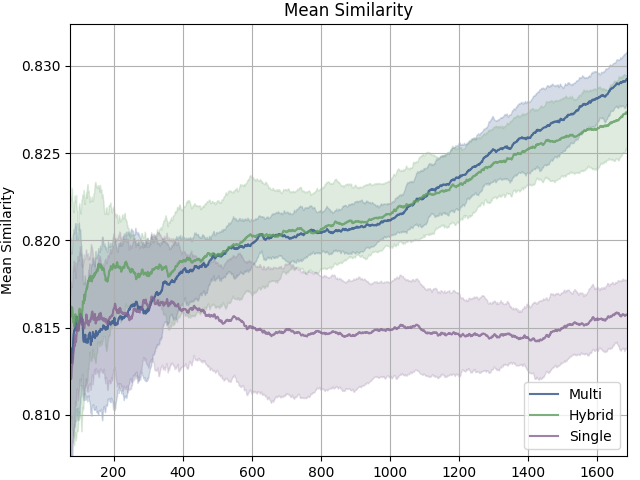}}\quad
  \caption{Smoothed mean similarity per training episode. Each graph is conscripted of 8 training sessions per case and then averaged and plotted with its standard deviation. For similar tasks, the similarity metric indicates effective training of the actor policy network. For an increasing number of agents, the similarity for single-critic networks decreases in relation to the other architectures (Best viewed in color). }
\end{figure}

\begin{figure}[]
  \centering
  \subfigure[n=2 agents
  ]{\includegraphics[scale=0.25]{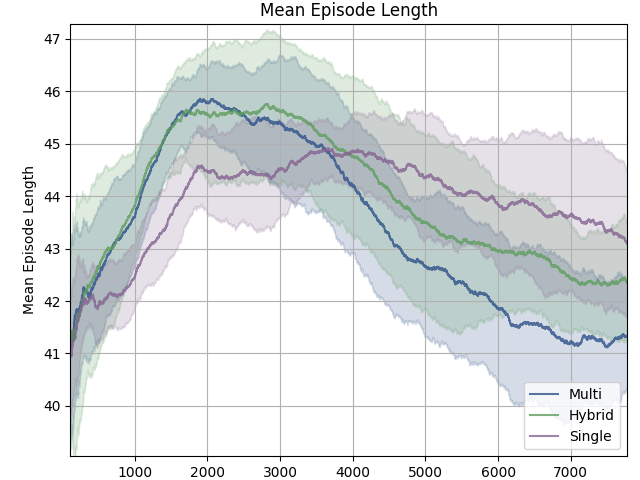}}\quad
  \subfigure[n=3 agents
  ]{\includegraphics[scale=0.25]{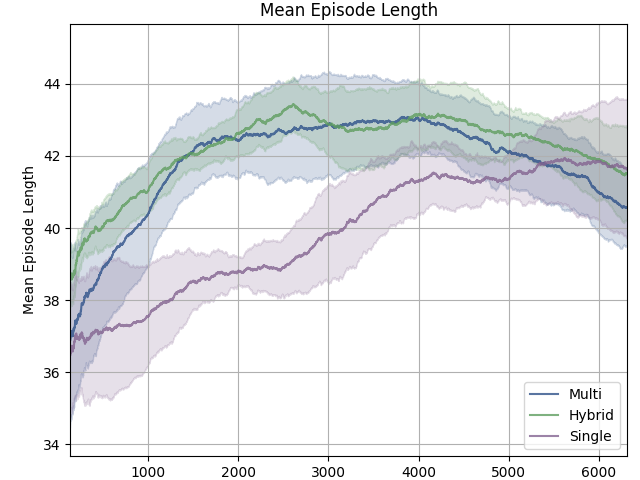}}\quad
  \subfigure[n=4 agents
  ]{\includegraphics[scale=0.25]{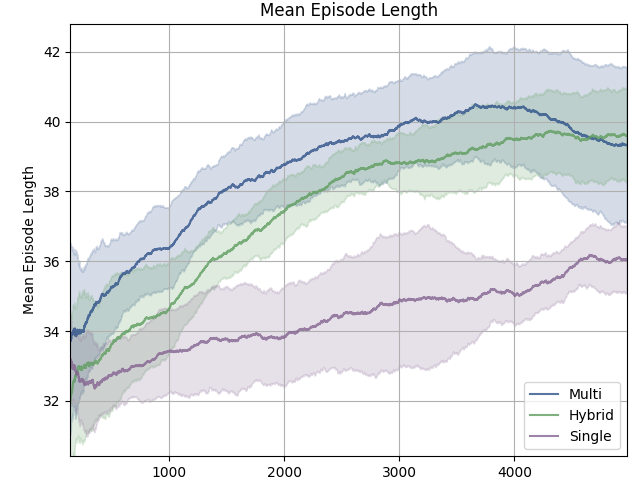}}\quad
  \subfigure[n=5 agents
  ]{\includegraphics[scale=0.25]{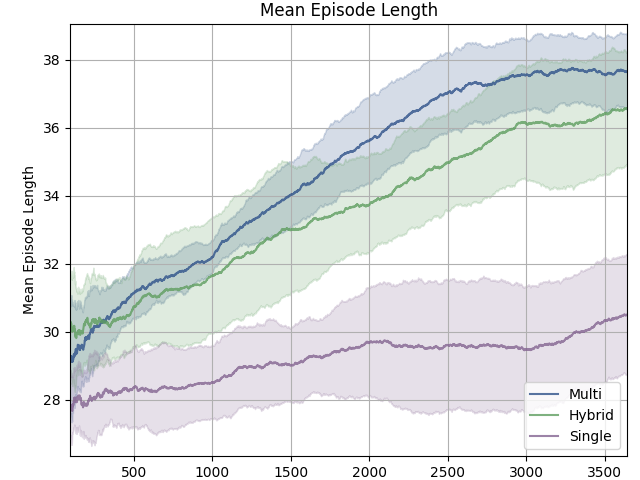}}\quad
  \subfigure[n=6 agents
  ]{\includegraphics[scale=0.25]{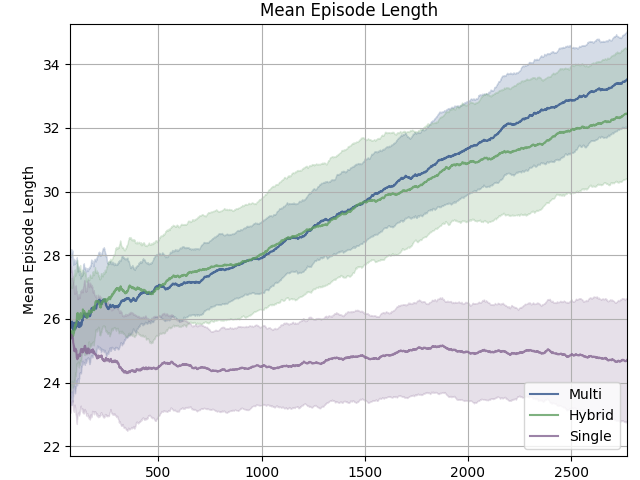}}\quad
  \subfigure[n=7 agents
  ]{\includegraphics[scale=0.25]{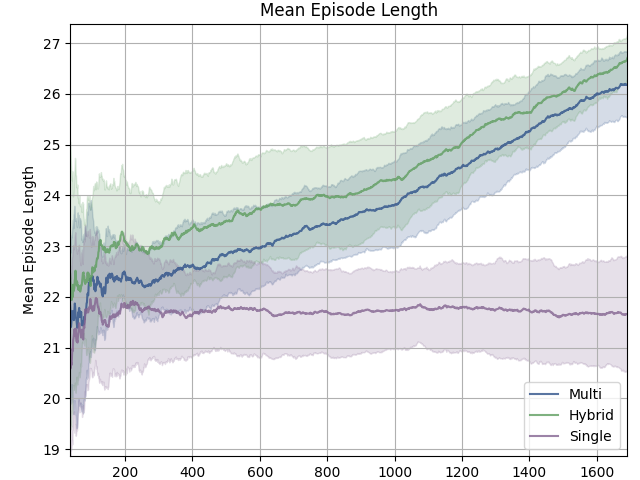}}\quad
  \caption{Smoothed mean episode length per training episode. Each graph is conscripted of 8 training sessions per case and then averaged and plotted with its standard deviation (Best viewed in color). }
\end{figure}
\subsection{Technical implementation}
The optimization algorithm was realized using PyTorch due to a better memory performance but a Tensorflow 2 implementation is also available in the repository of the code publication. Tensorboard event files extract a number of relevant metrics during training that we will later evaluate. Depending on the environment, the selected relevant metrics for performance evaluation include the absolute reward signal per episode, episode length and the customized similarity metric. 

The training sessions were performed on Alice2, part of the High-performance computing (HPC) Cluster of the University of Leicester. Simulations were run simultaneously. Usually, 8 training sessions of 150K episodes were performed per agent number and architecture for each environment. The mean metrics and their respective standard deviations form significant performance evaluation for the mentioned architectures. It has to be mentioned that the episode axis in the diagrams does not represent the absolute training episode but the relative episode number. Generally, all the environments are trained with a constant episode number of 150K episodes. Due to a constant buffer size of a set of observations and various episode length due to abort conditions that indirectly depend on the number of agents, the episode labels are relative.

Together with other environment instances available in the published code, we introduce a simple targeting task where multiple drones have to approach a target without collision. The environment extracts multiple reward signals and a single actor policy network extracts an action vector that is split and distributed to the drones. Respective multiple reward signals are then returned. 

\begin{figure*}[]
  \subfigure[
  ]{\includegraphics[scale=0.18]{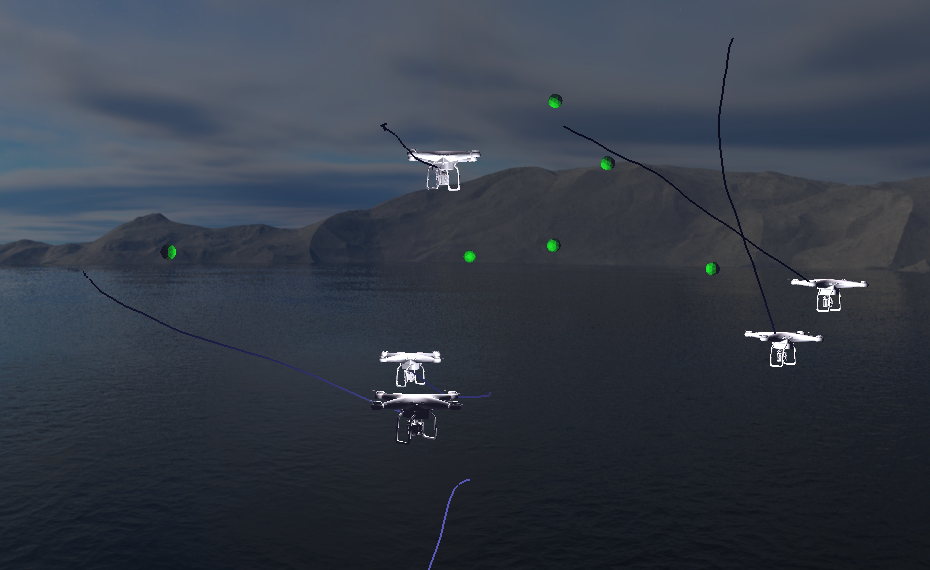}}\quad
  \subfigure[
  ]{\includegraphics[scale=0.18]{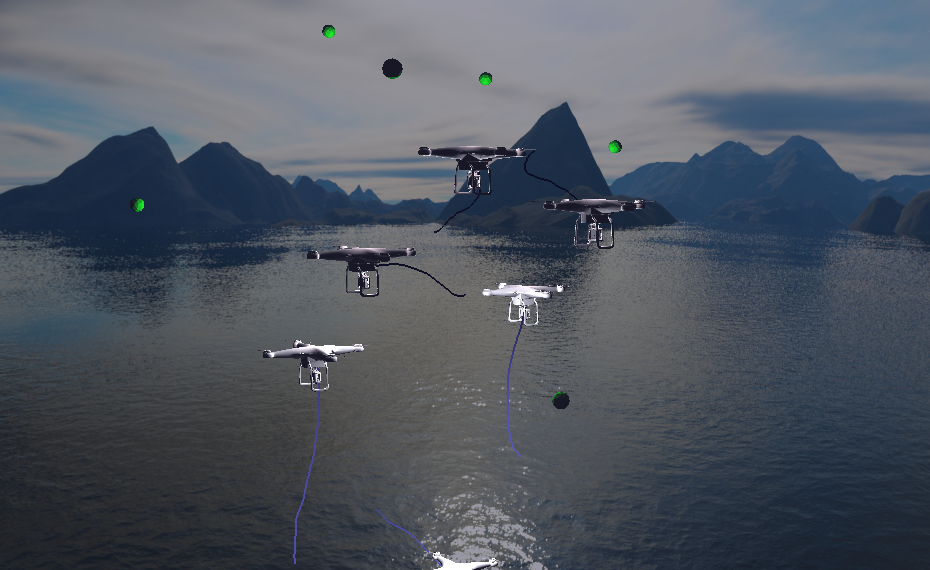}}\quad
  \subfigure[
  ]{\includegraphics[scale=0.18]{images/3.png}}\quad
  \subfigure[
  ]{\includegraphics[scale=0.18]{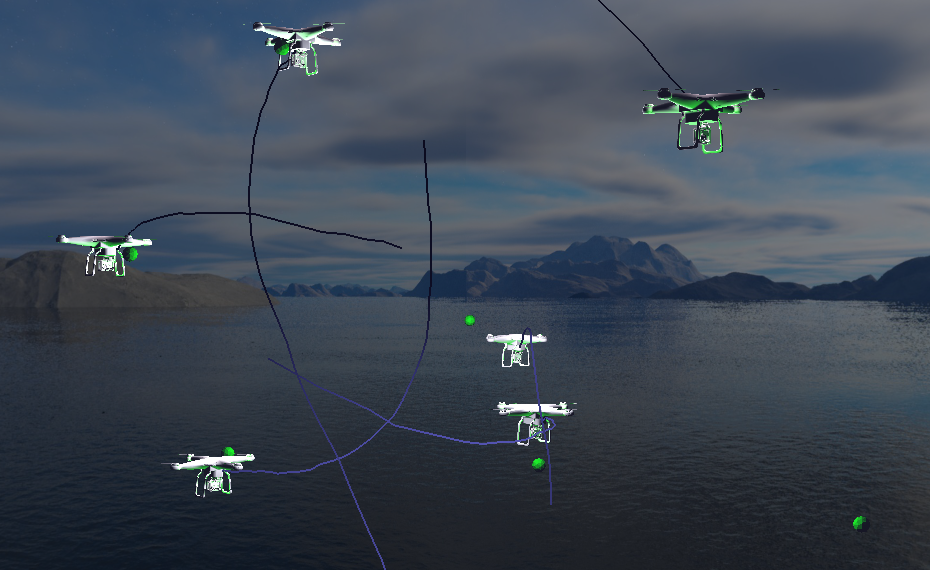}}\quad
  \subfigure[
  ]{\includegraphics[scale=0.18]{images/5.png}}\quad
  \subfigure[
  ]{\includegraphics[scale=0.18]{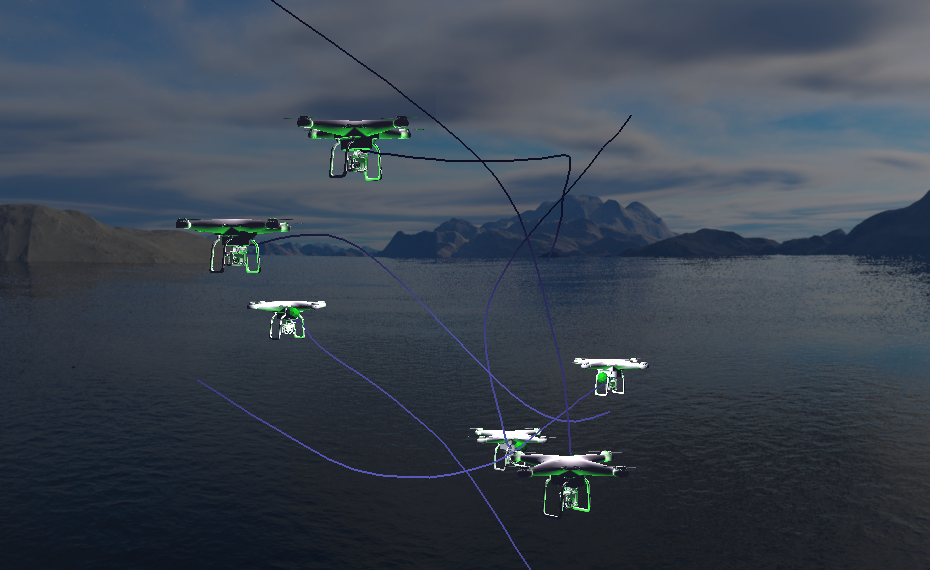}}\quad
  \caption{For the multi-agent coordination task in a physics-based reinforcement learning environment, we observe the improvement of actor-policy, including the ability to coordinate and plan trajectories being aware of possible collisions.  (a) An early training episode with nearly random non-normalized action space results in little deviation from the current momentum. (b) Actor policy is improving by starting to approach the states of higher value estimation, concurrently value estimation is becoming more accurate. (c) Actor policy is gaining an ability to approach the targets with little collision awareness and do not slow down before reaching a target. (d) Better speed adaption as approaching objectives, improved ability to avoid collisions. (e) Collision awareness with the expense of optimal path planning, but with an enhanced ability to control the speed. (f) High level of coordination, achieving nearly optimal trajectories and speeds whilst avoiding collisions. It becomes apparent that with increasing episodes, the ability to adapt speeds to approach a target or avoid collision becomes more sophisticated, and as a result, the trajectories are similar according to the principle of the last action. The model-free environment creates new initial positions for targets and drones at the beginning of each episode. Training is performed without rendering, but at regular distances, traces are extracted and subsequently rendered. The mean episode length indicates the number of collisions (Best viewed in color). }
\end{figure*}

\subsection{Metrics}
To measure the asymmetry in sub-reward optimization, a metric is necessary to reflect this asymmetry. To determine a metric, we consider each sub-objective as a dimension and calculate the center of mass in $\mathbb{R}^n$ as $(r_1, r_2, ..., r_n)$. Then, each point is a reward extraction for a state depending on $t$ $(r_1(s_t), r_2(s_t), ..., r_n(s_t))$ for each $t$ in a defined batch size. The center of mass for this objective space is:
\begin{equation}
\hat{r} = \frac{1}{n} \sum_t^n (r_1(s_t),r_2(s_t),...,r_n(s_t))
\end{equation}
Inspired by cosine similarity, we determine the angle between our center of mass for the multi-dimensional objective space and the expected optimal distribution center of mass given as the normal vector: (1,1,..,1). With ideal similarity $\theta = 0$:
\begin{equation}
cos(\theta) = \frac{\hat{r} \cdot \textbf{1}}{|\hat{r}| \cdot |\textbf{1}|}
\end{equation}
Our normalized similarity with an ideal value of 1.0 is therefore obtained with: 
\begin{equation}
S(\hat{r}) = 1- |1-\frac{\hat{r} \cdot \textbf{1}}{|\hat{r}| \cdot |\textbf{1}|}|
\end{equation}

\subsection{Observations}
For this environment, we observe that standard single-critic architectures fail to optimally train an efficient actor policy for multiple agents. When comparing the rewards, it can be observed that the policy optimizes the dominant rewards. Hybrid architectures, based on a single-critic that extracts multiple value estimations, typically perform better than single-critic networks but are generally outperformed by multi-critic polymorph architectures.

A general observation is that for a small number of agents (e.g. $n$=2), the difference in performance is nearly neglectable. When the advantage function only depends on a small number of agents, a state value estimation is relatively reliable but for a higher number of agents, the performance gap becomes apparent. 

One may argue that for architectures based on a single critic with multiple value extraction, the comparison against polymorph architectures is not applicable since the number of neurons differs. In essence, the practical advantage of having multiple critics is the ability to instantiate and destroy networks dynamically. Corresponding to the agents that terminate before the episode end or are newly created dynamically, their respective value estimation network does not need to be trained anymore. In such cases, a constant size value estimation network may even destroy the previously established advance by modifying the established weights. An observation can be confirmed in the experiments. 

As the most relevant metric for performance evaluation, we tabulate the mean aggregated reward signal in Table I. The final value for a constant number of 150K episodes together with the development over training episodes shown in Figure 8 gives a clear image for various constellations with a changing number of agents. In the case of $n$=2, the examined architectures seem to perform nearly identical. The ability of single-critic and hybrid architectures to derive an advantage out of the two values, or in other words, a majority voting for two critics seems to work well. Single-critic networks perform worse for an increasing number of agents. It seems the development stops with a certain number of agents and the system performance seems to deteriorate. As expected, the single value estimating critic fails to trace back the origin of the gradient change for a higher number of agents. For hybrid networks, based on a single critic that extracts multiple value estimations, the observations seem diverse. The performance in terms of mean aggregated reward signals is similar but slightly worse than the one of multi-critic networks. The gradient can trace back the changes needed to create a successful policy. Still, in the cases that require dynamic instantiating of agents and the agents that terminate before an episode finishes, hybrid networks are doomed to impair the performance of other agents due to the update of weights. In multi-critic networks, we separate the networks to avoid the impairment of system performance in case the number of agents changes within an episode. 

As a symptomatic issue of multiple reward signal reinforcement learning, different update speeds for reward sources can be analyzed using the introduced multi-dimensional similarity metric as constituted in Table II and the episodic development shown in Figure 9. Again, the single-critic networks demonstrate the worst performance with an increasing number of agents. Similarly, the mean similarity per training episode performs best for multi-critic architectures and slightly worse for hybrid networks. 

Finally, the selected task of cooperation of multiple drones is evaluated on its cooperation ability. Since the collision of drones early terminates the episode, the metric of smoothed mean episode length per training episode allows us to judge the ability of cooperation. Here again, single-critic networks perform worse with an increasing number of agents, and the multi-critic networks perform better than the hybrid networks. Objectively, this performance difference will be more evident for dynamic applications. 

For all the examined environments based on the base environments for UAV coordination, we have the same conclusions, and the experimental results will be part of the published repository.

\subsection{Convergence and complexity analysis}

The convergence analysis is complex. We will discuss the convergence to an optimal actor policy. For our architecture, the convergence to an optimal policy depends also on the neural network design and its depth. The figure showing the accumulated reward signal development shows the characteristic convergence at a stage where rewards seem to remain constant. That is not achievable at all for single-critic networks with more than 3 or 4 agents, depending on the engaged environment. Training breaks down, and no improvement or even deterioration has been observed. Usually, control problems based on the positional states can be solved using fully connected layers. The convergence depends on the previous states. The Markov assumption of dependency on the neighboring state is not necessarily relevant for some of the control problems presented. 



\subsection{Additional experiments}

Additionally, we performed a significant number of experiments using customized environments that emphasize both cases, a multi-task configuration with non-conflicting split actions and concatenated states, and a multi-objective configuration with conflicting objectives and exclusive action vectors. The experiments were performed with various numbers $n$ of objectives. For this purpose, we customized environments with continuous actions for multiple objectives. Based on the OpenAi gym environments \cite{OpenAiGym}, we created environments that simulate both multi-task and multi-objective reinforcement learning. Inspired by those environments, we created customized environments simulating a robotic arm to learn a choreography. For the case of multi-task environments, the task is simply duplicated (an environment duplication can be achieved for standard environments of \cite{OpenAiGym}). For the conflicting case, multiple sub-targets are initiated to obtain a Pareto optimum. Our customized environment ArmEnv extracts multiple reward signals. The metric of similarity as initially used to derive a loss to minimize the pace difference is used again to observe the development of different reward signals. We observed similar performance advantages for the above-mentioned cases, achieved by the previously discussed multi-agent environments.


\begin{figure}[]
  \subfigure[
  ]{\includegraphics[height=2.4cm]{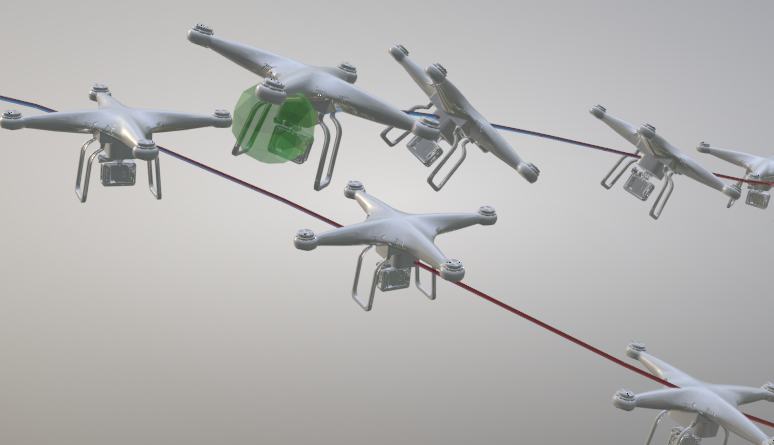}}\quad
  \subfigure[
  ]{\includegraphics[height=2.4cm]{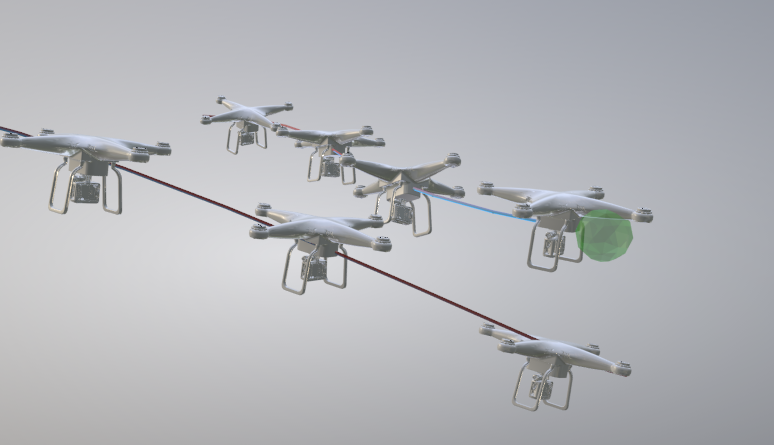}}\quad
  \subfigure[
  ]{\includegraphics[height=2.4cm]{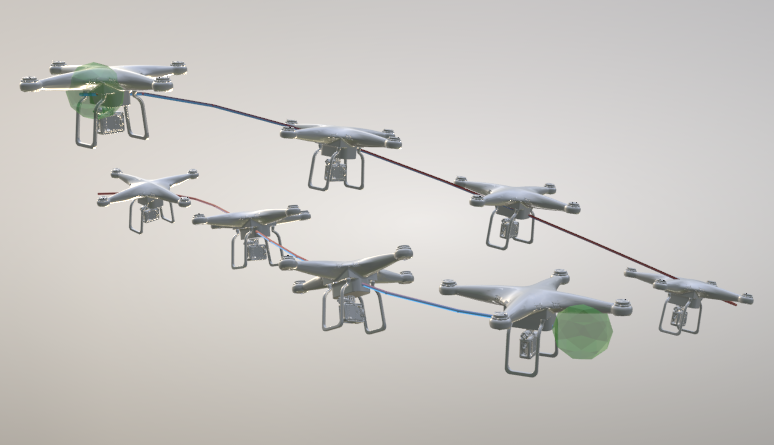}}\quad
  \subfigure[
  ]{\includegraphics[height=2.4cm]{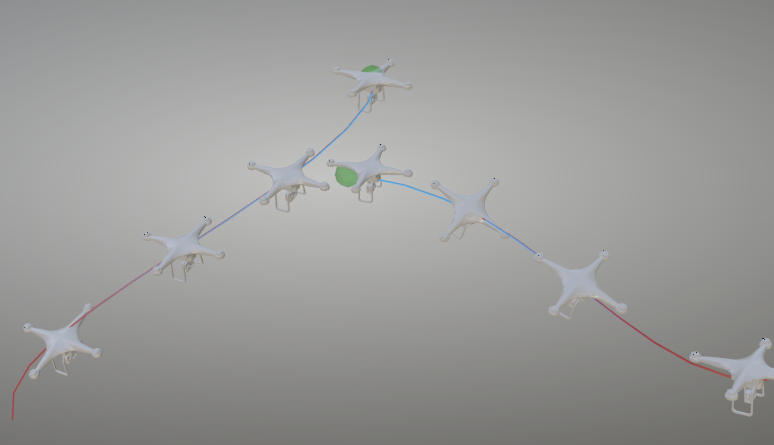}}\quad
  \caption{Visualization of adaption of speed from blue (slow) to red (fast) and the learned agility by tilt and orientation adaption that both serve to avoid collisions and achieve optimal coordination (Best viewed in color). }
\end{figure}

\section{Conclusions}

We have devised a methodology to control groups of Unmanned Aerial Vehicles using policy gradient optimization based on a single actor policy. Multiple value estimation networks of different depths and design can enable an advantage function to optimize a stochastic actor policy for optimal coordination of agents. Our experiments confirm that policy-gradient based multi-agent reinforcement learning can be applied for a higher number of agents and that a high level of coordination can be achieved using a single actor policy network. Using several physics-based reinforcement learning environments for multiple agents with various coordination tasks, we have achieved a high symmetry in reward signal development, that indicates an effective use of actor-network neurons, and the avoidance of duplication for similar tasks. Further experimental investigations in the highlighted applications of Multi-Objective Optimization and Multi-Objective Reinforcement Learning might show promising performance of Poliymorph Policy Gradient Optimization. 


\nocite{Multi-Quadcopter-env, Quadcopter-env, MORLGeneralized, ScalarizedMORL, MultiTaskLearning, MultiTaskRL, MultiAgentBook, MultiAgentCooperation, MultiAgentOverview, MultiAgentPartially, MultiAgentRL, Tzeng, MORLGeneralized, WeightedSum, EpsilonConstraints, MORLGeneralized, ProblemsAISafety, MultiTaskRL, A2CA3C, policyGradientsLilianWeng, DeepQPartially, SelfLearningQ, DeepQPartially, StanfordPaper}


\bibliographystyle{IEEEtran}
\bibliography{refs}


\begin{IEEEbiographynophoto}{Yoav Alon}
received a Master of Science in Computer Science at the University of Leicester and is currently a PhD-Student at the School of Informatics of the University of Leicester. 
\end{IEEEbiographynophoto}
\begin{IEEEbiographynophoto}{Huiyu Zhou}
received a Bachelor of Engineering degree in Radio Technology from Huazhong University of Science and Technology of China and a Master of Science degree in Biomedical Engineering from University of Dundee of United Kingdom, respectively. He was awarded a Doctor of Philosophy degree in Computer Vision from Heriot-Watt University, Edinburgh, United Kingdom. Prof. Zhou currently heads the Applied Algorithms and AI (AAAI) Theme and leads the Biomedical Image Processing Lab at University of Leicester. Prior to this appointment, he worked as a Lecturer (2012-17) at the School of Electronics, Electrical Engineering and Computer Science, Queen's University Belfast (QUB). 
\end{IEEEbiographynophoto}

\section{Supplementary material}
The Github code repositories for this work can be found at \cite{Multi-Quadcopter-env} and \cite{Quadcopter-env}.

\begin{figure*}[]
  \subfigure[Single-Critic for n = 2
  ]{\includegraphics[scale=0.18]{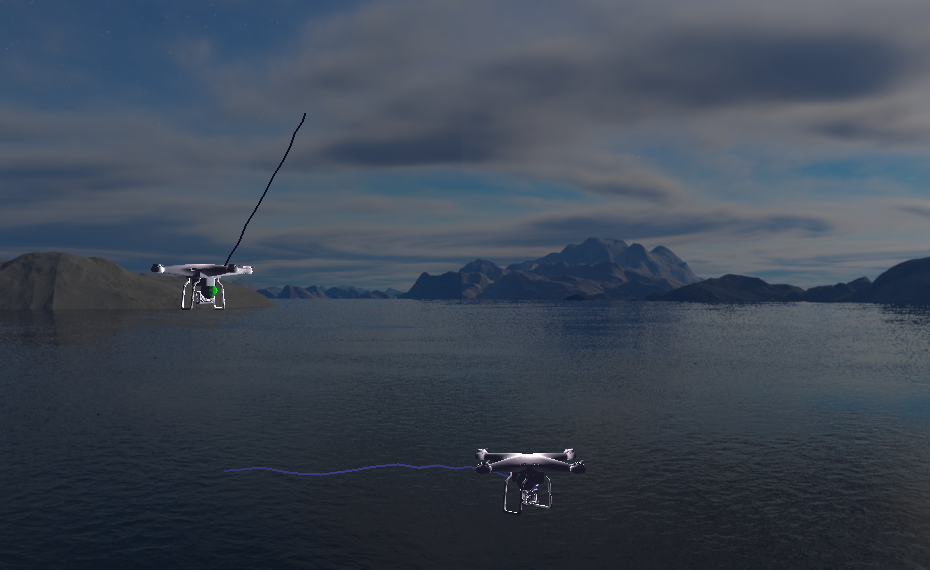}}\quad
  \subfigure[Sincle-Critic for n = 3
  ]{\includegraphics[scale=0.18]{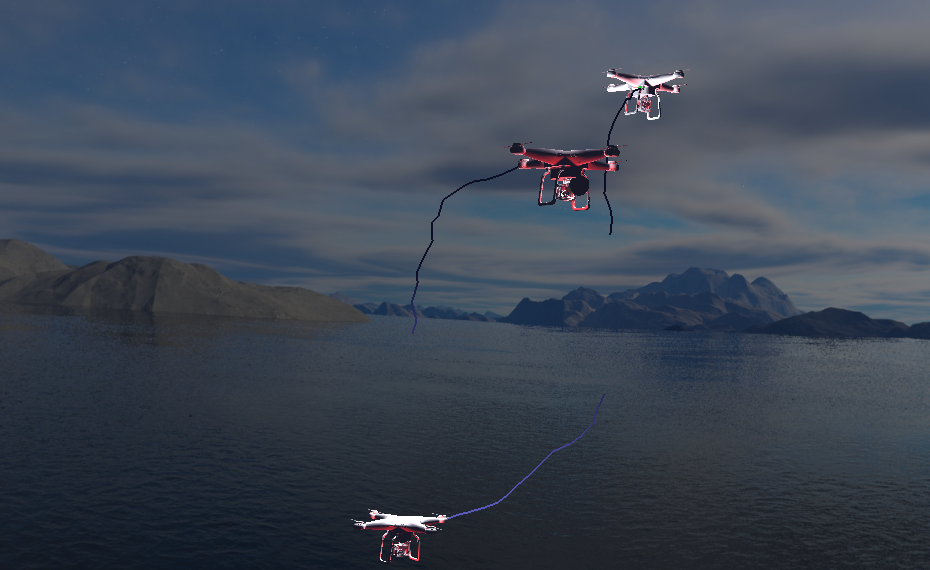}}\quad
  \subfigure[Single-Critic for n = 4
  ]{\includegraphics[scale=0.18]{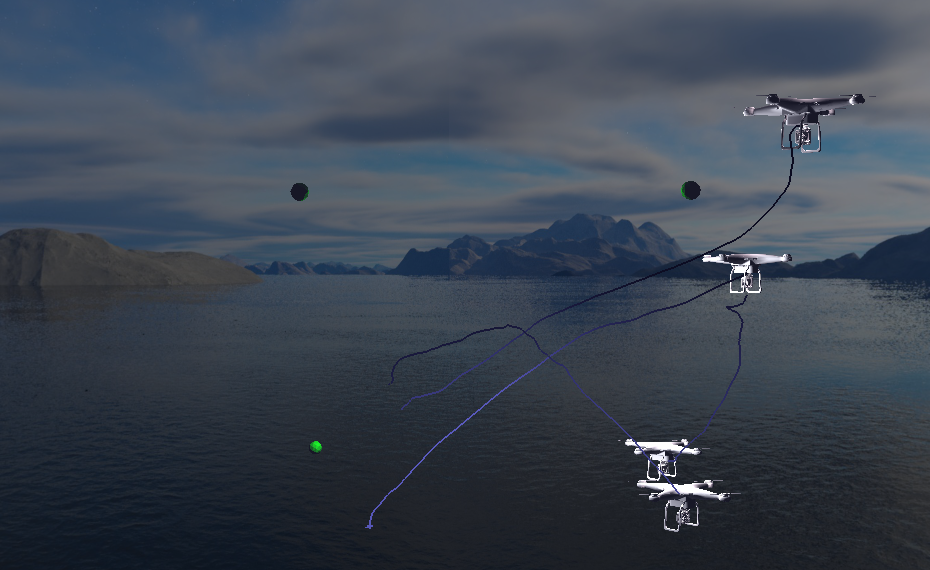}}\quad
  \subfigure[Multi-Critic for n = 2
  ]{\includegraphics[scale=0.18]{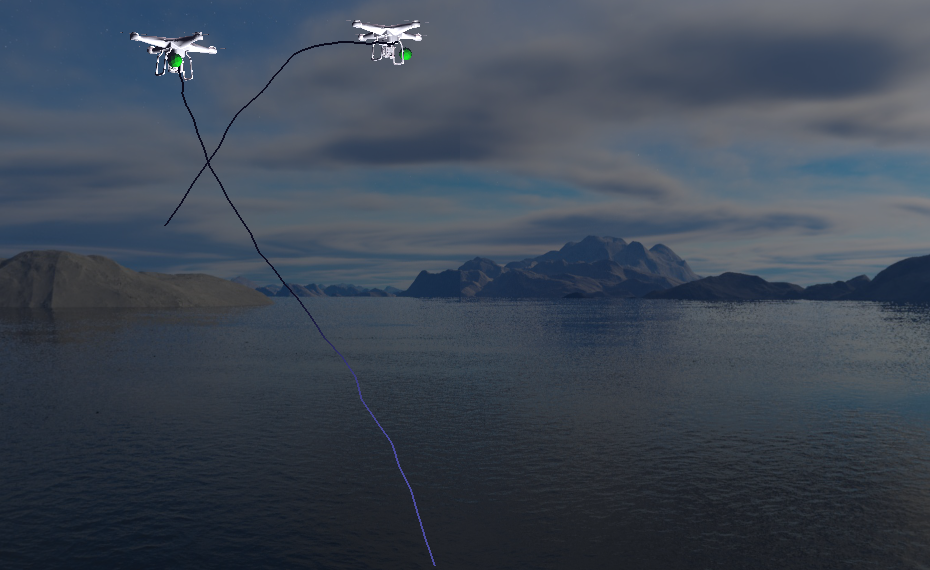}}\quad
  \subfigure[Multi-Critic for n = 3
  ]{\includegraphics[scale=0.18]{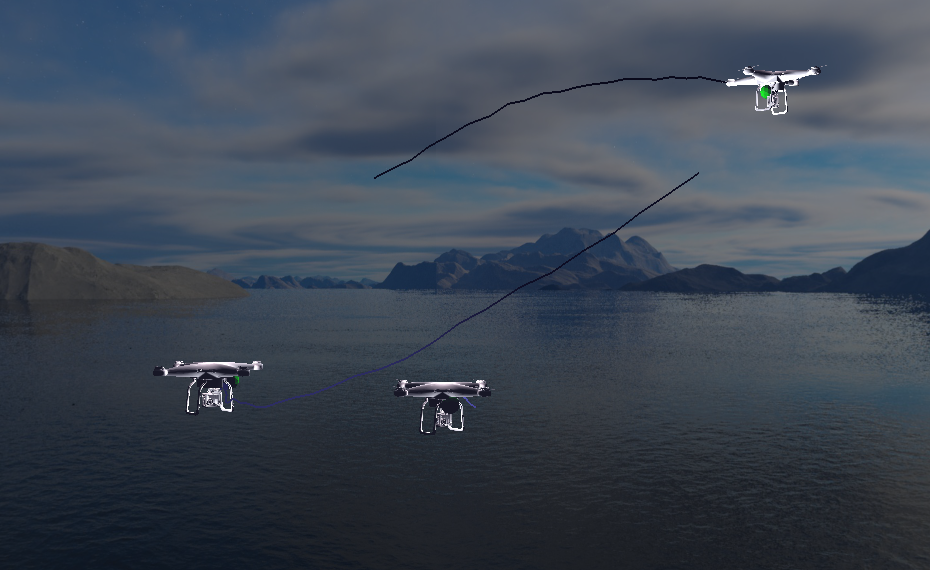}}\quad
  \subfigure[Multi-Critic for n = 4
  ]{\includegraphics[scale=0.18]{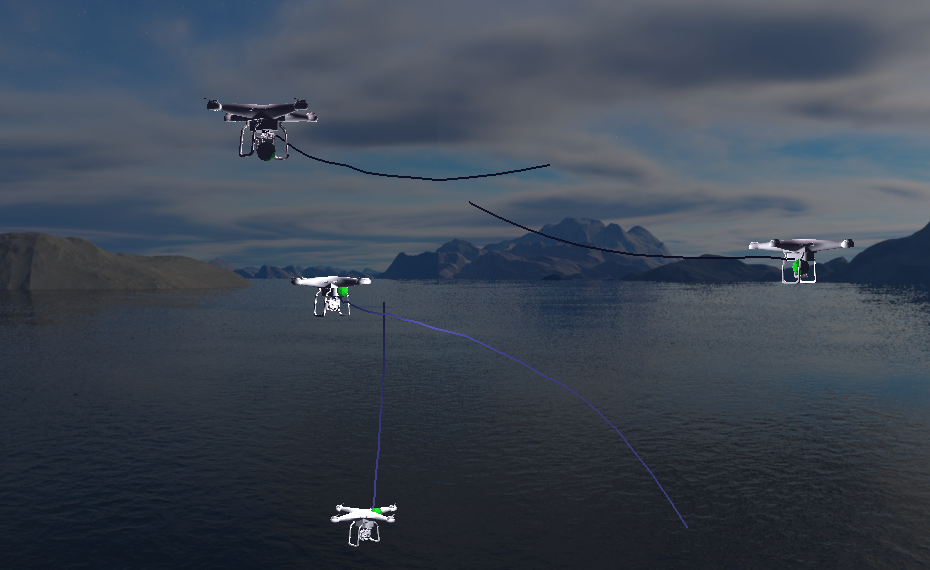}}\quad
  \caption{For a relatively small number of agents, the performance of single-critic and multi-critic architectures are similar, even if the point of convergence is delayed to a later episode. However, for an increasing number of agents, the optimizer of single-critic architectures fails to train a converging actor policy, and the performance even deteriorates (Best viewed in color). }
\end{figure*}

\begin{figure*}[]
  \subfigure[Single-Critic for n = 5
  ]{\includegraphics[scale=0.18]{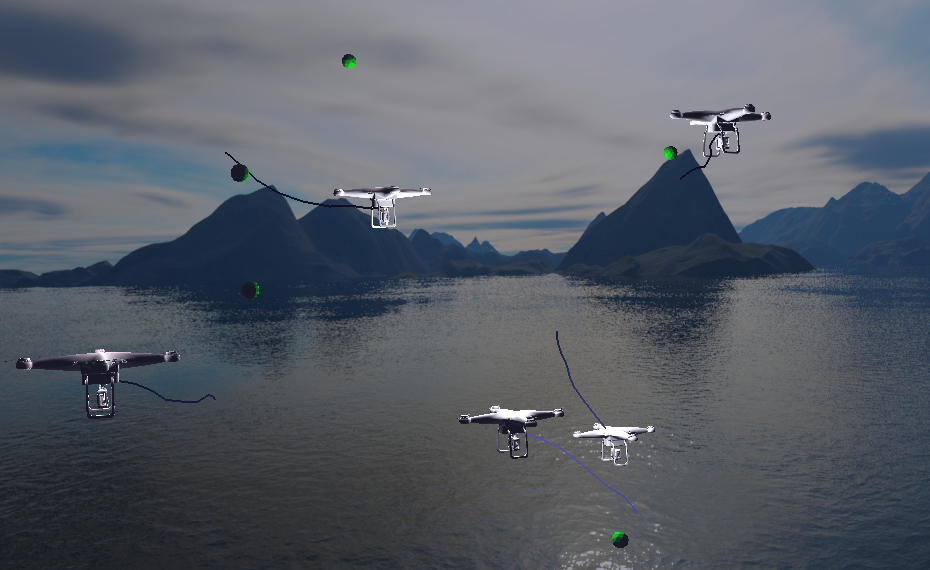}}\quad
  \subfigure[Single-Critic for n = 6
  ]{\includegraphics[scale=0.18]{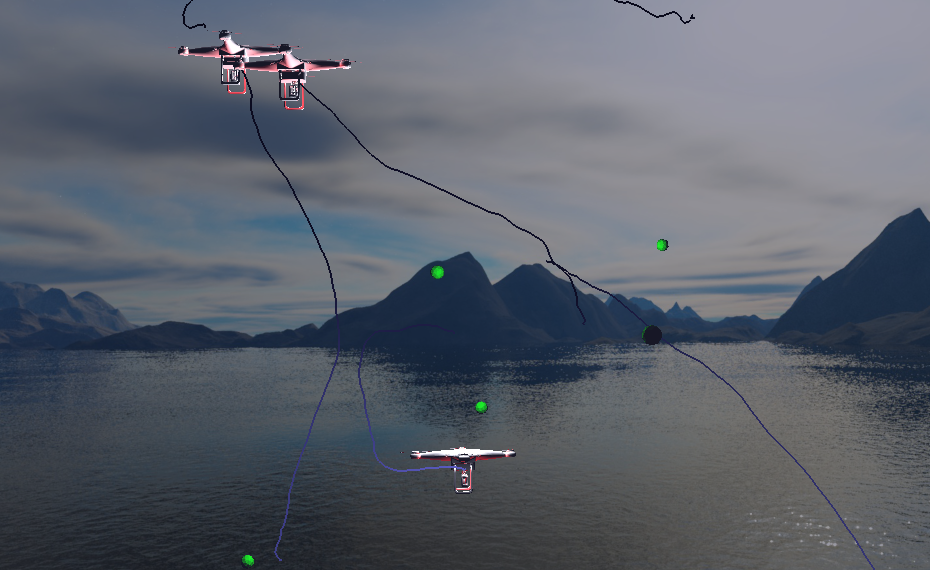}}\quad
  \subfigure[Single-Critic for n = 12
  ]{\includegraphics[scale=0.18]{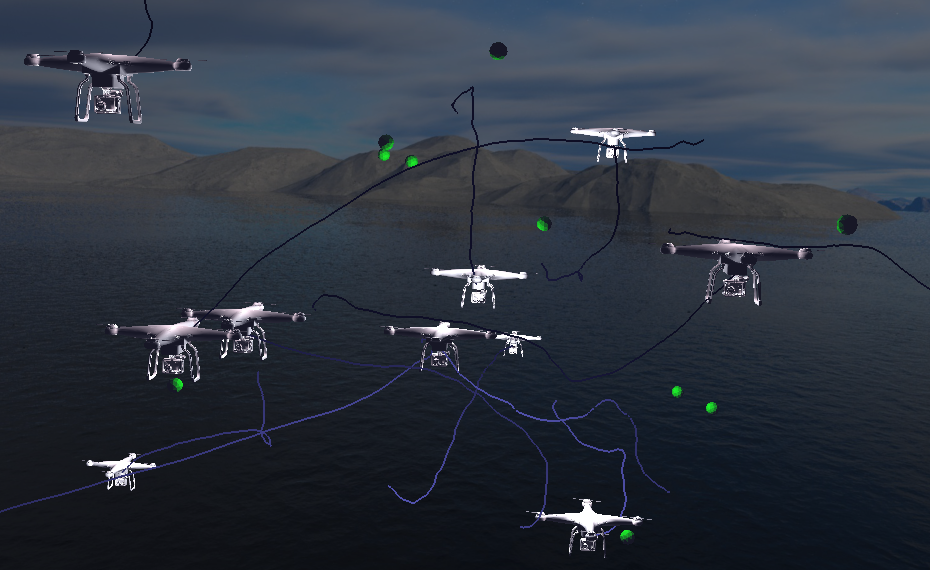}}\quad
  \subfigure[Multi-Critic for n = 5
  ]{\includegraphics[scale=0.18]{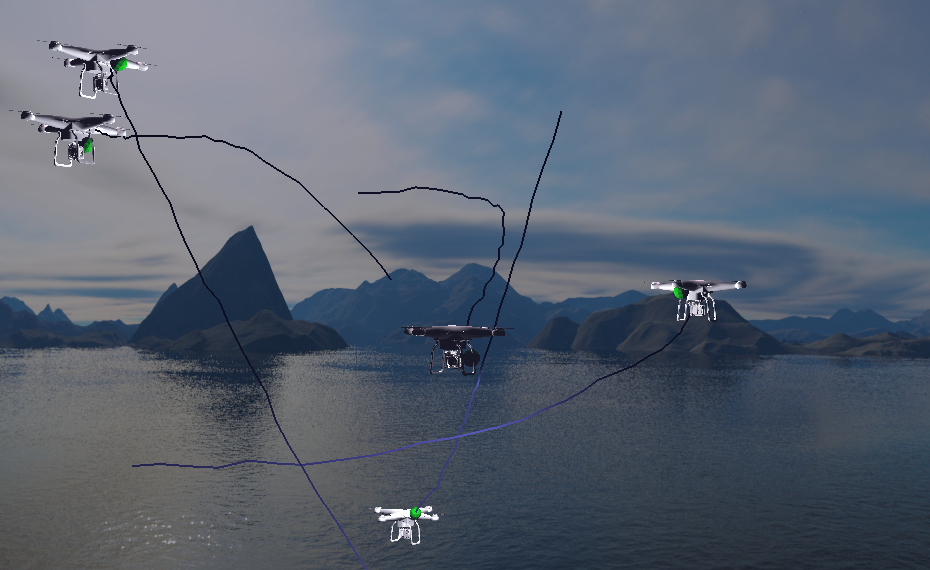}}\quad
  \subfigure[Multi-Critic for n = 6
  ]{\includegraphics[scale=0.18]{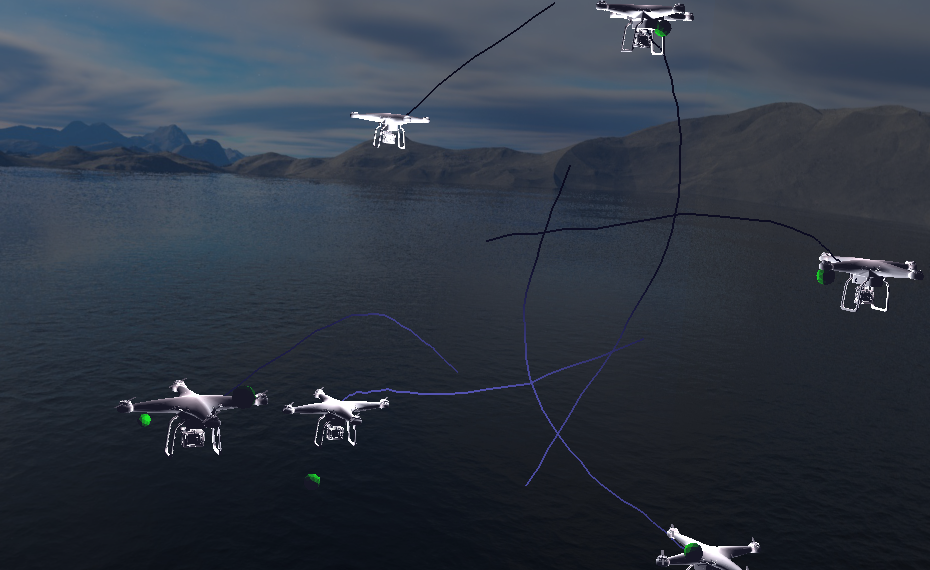}}\quad
  \subfigure[Multi-Critic for n = 12
  ]{\includegraphics[scale=0.18]{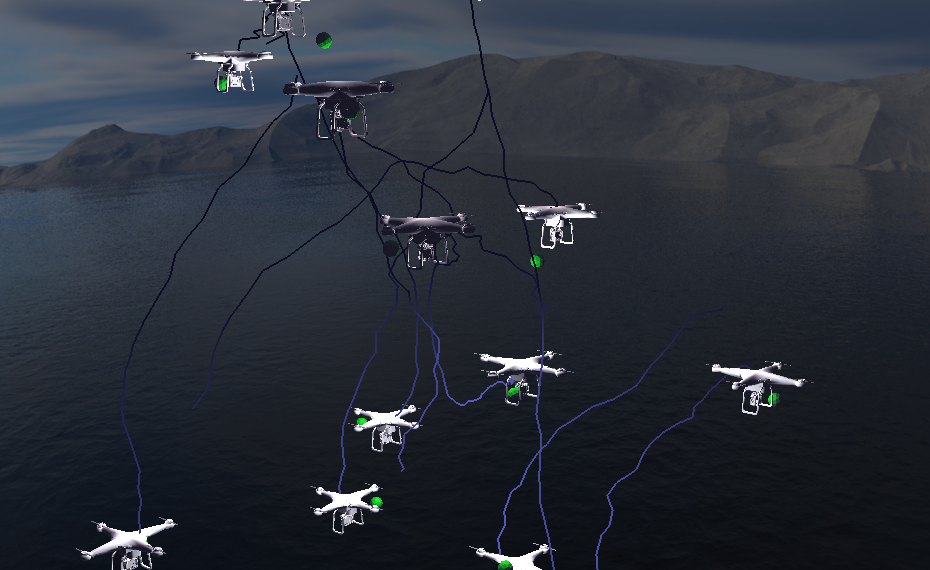}}\quad
  \caption{For an increasing number of agents, the inability to train single-critic architectures becomes evident. This  results in a seemingly random policy and collisions as shown in (b) and (c) (Best viewed in color). }
\end{figure*}

\begin{figure*}[]
  \subfigure[n = 4, Single-Critic
  ]{\includegraphics[scale=0.18]{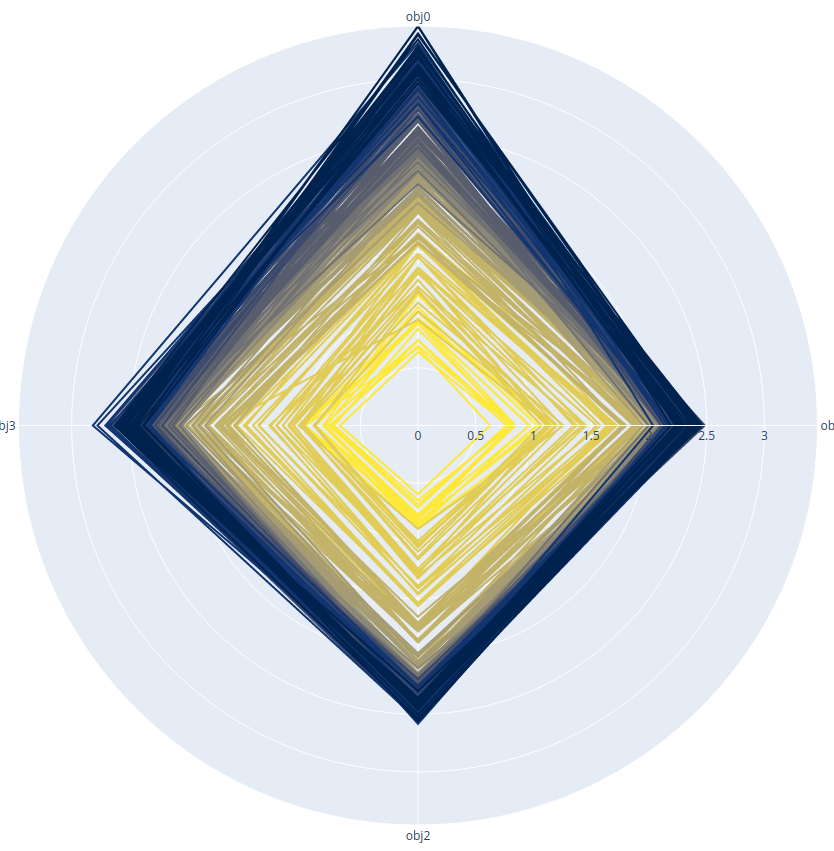}}\quad
  \subfigure[n = 6, Single-Critic
  ]{\includegraphics[scale=0.18]{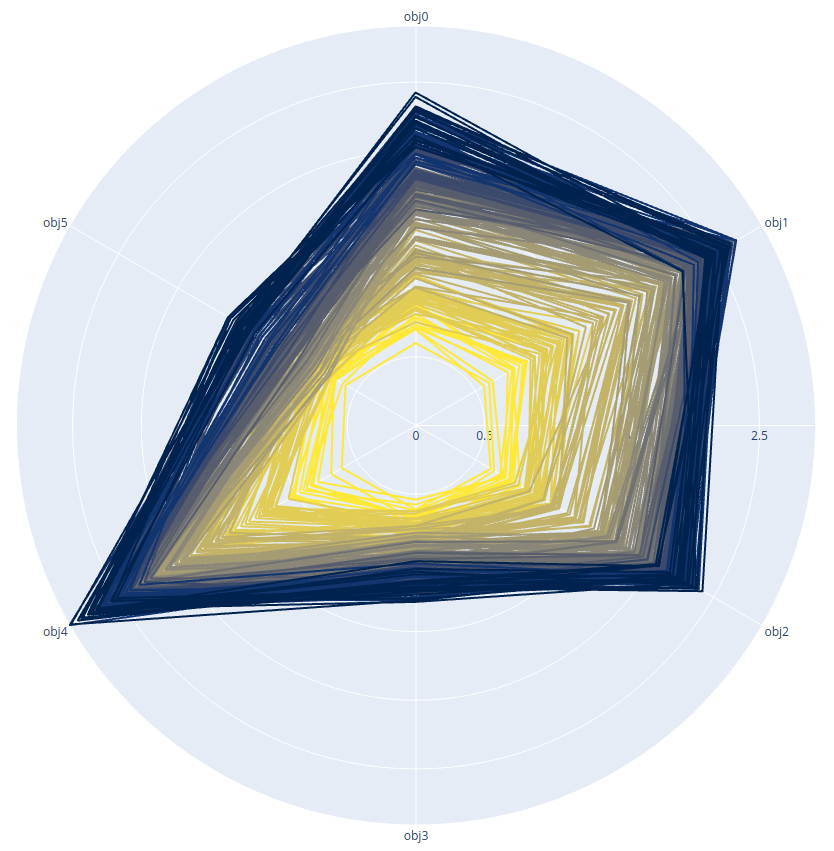}}\quad
  \subfigure[n = 8, Single-Critic
  ]{\includegraphics[scale=0.18]{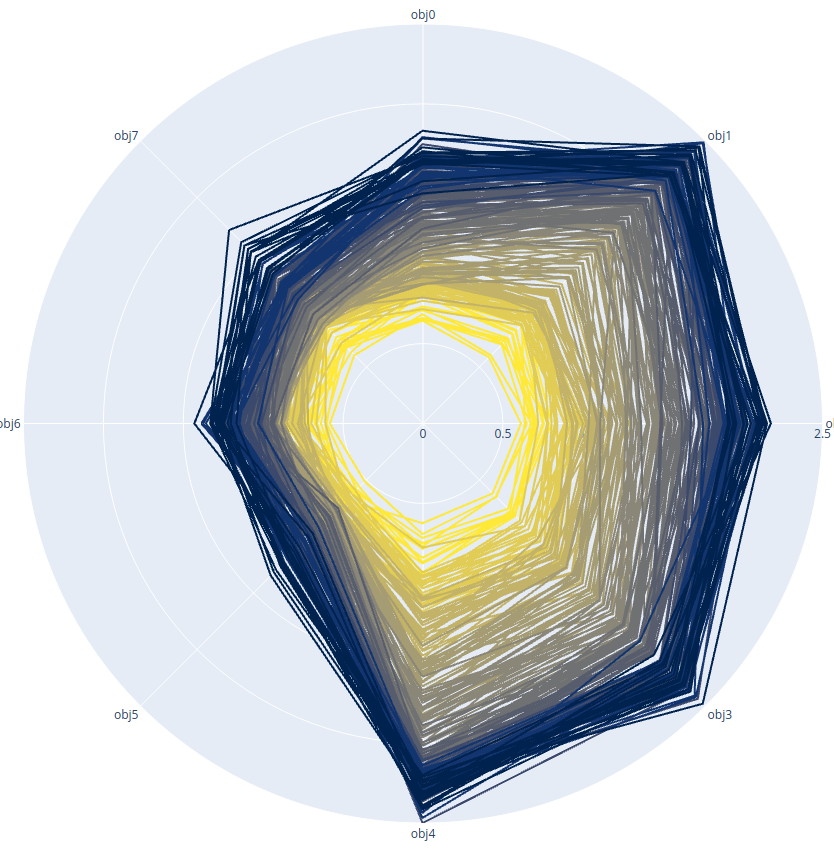}}\quad
  \subfigure[n = 4, Multi-Critic
  ]{\includegraphics[scale=0.18]{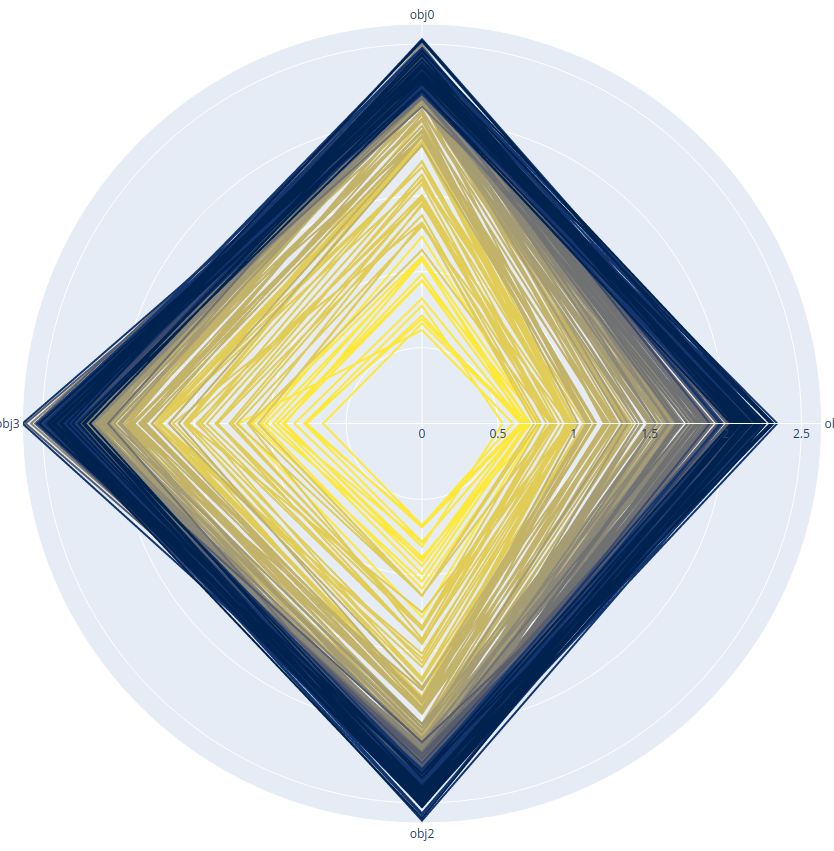}}\quad
  \subfigure[n = 6, Multi-Critic
  ]{\includegraphics[scale=0.18]{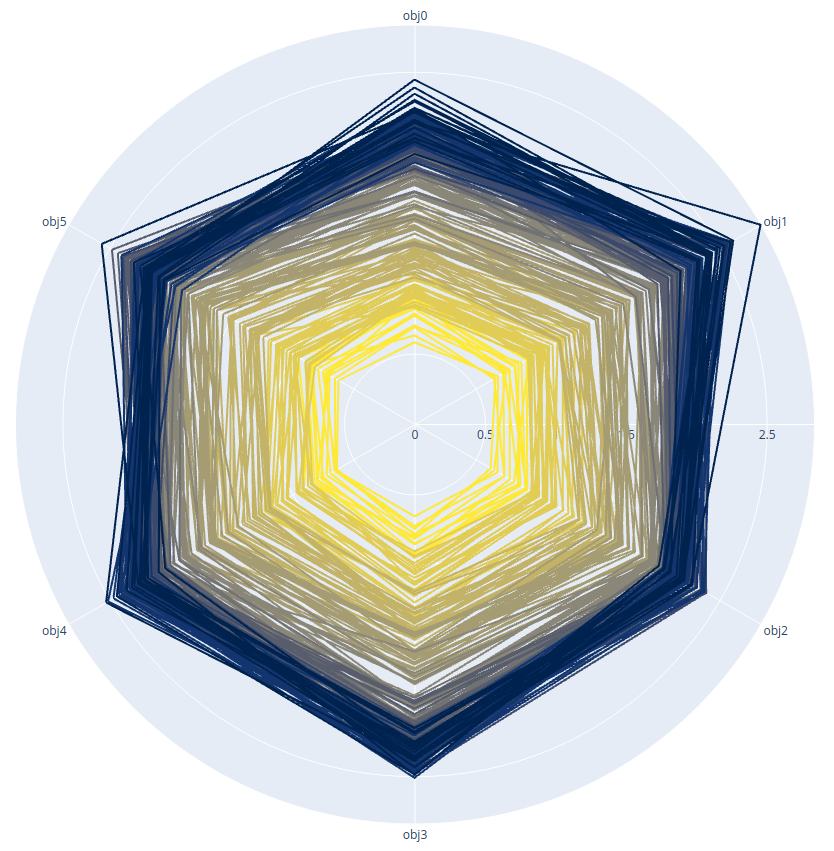}}\quad
  \subfigure[n = 8, Multi-Critic
  ]{\includegraphics[scale=0.18]{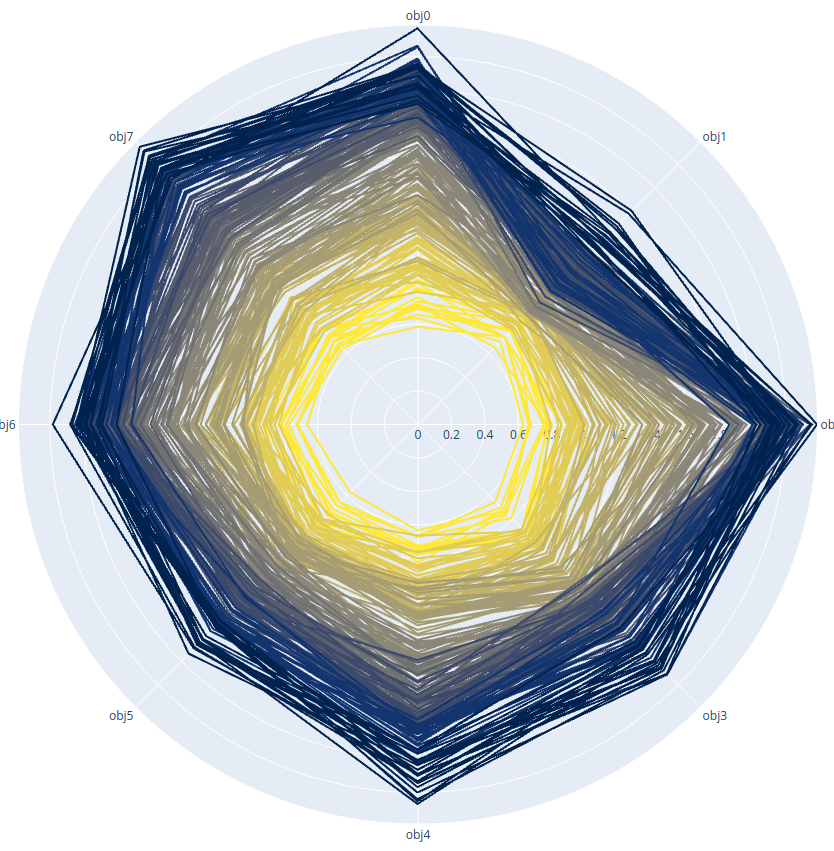}}\quad
  \caption{A number of $n$ non-conflicting objectives using a single critic network architecture (a-c) and a multi-critic architecture (d-f). The high symmetry of sub-figures (d)-(f) for multi-critic policy gradient optimization is evident in this example, showing better optimization for each objective. This example seems that the symmetry for multi-critic optimization decreases with a higher number of objectives. In a real scenario, the number of the neurons will increase for a higher number of the objectives (Best viewed in color). }
\end{figure*}


\begin{table*}[]
\caption{The aggregated reward signals for multiple agents.  }
\begin{tabular}{|p{1cm} |p{1cm}||p{3cm}||p{1.5cm}|p{1.5cm}| p{1.5cm}|p{1.5cm}|p{3cm}|}
 \hline
 \multicolumn{7}{|c|}{Aggregated reward signals} \\
 \hline
 Agents& Method &Mean&$\sigma$& $R_{min}$ & $R_{max}$ & Training iterations\\
 \hline
 2 & multi  & \textbf{68.58} & 3.38 & 32.33 & 72.93 & 7956.0   \\ 
 & hybrid & 66.73 & 3.65 & 15.32 & 70.78 & 9718.125 \\ 
 & single & 65.07 & 5.24 & 31.86 & 73.86 & 9276.625 \\ \hline
3 & multi  & 59.08 & 4.96 & 42.81 & 77.09 & 6514.25  \\ 
 & hybrid & \textbf{62.46} & 4.81 & 39.45 & 67.64 & 8259.25  \\ 
 & single & 56.64 & 4.42 & 34.73 & 70.13 & 7413.625 \\ \hline
4 & multi  & \textbf{57.08} & 5.83 & 33.53 & 73.8  & 5429.75  \\ 
 & hybrid & 51.76 & 3.7  & 8.31  & 57.54 & 6452.375 \\
 & single & 46.71 & 2.51 & 40.52 & 69.02 & 5338.125 \\ \hline
5 & multi  & \textbf{49.14} & 4.66 & 16.84 & 55.31 & 4609.625 \\ 
 & hybrid & 45.09 & 4.61 & 33.42 & 77.72 & 5200.75  \\ 
 & single & 41.38 & 1.93 & 29.33 & 54.75 & 4024.375 \\ \hline
6 & multi  & \textbf{43.51} & 4.03 & 6.63  & 49.56 & 3965.25  \\ 
 & hybrid & 37.04 & 2.79 & 33.6  & 79.55 & 4209.0   \\ 
 & single & 33.74 & 0.98 & 26.81 & 68.78 & 3124.25  \\ \hline
7 & multi  & 31.23 & 1.42 & 18.85 & 35.36 & 1753.75  \\ 
 & hybrid & \textbf{32.43} & 2.48 & 28.41 & 50.57 & 3284.125 \\ 
 & single & 29.33 & 1.69 & 7.44  & 32.11 & 2620.25  \\ \hline \hline
10 & multi & \textbf{22.34} & 1.23 & 11.08 & 26.89 & 1966.125 \\ 
 & hybrid & 22.16 & 1.03 & 14.68 & 29.98 & 1894.875 \\ 
 & single & 19.45 & 0.69 & 4.8 & 28.33 & 1695.5 \\ \hline 
14 & multi & \textbf{13.43} & 0.7 & 6.2 & 15.22 & 1212.625 \\ 
 & hybrid & 12.72 & 1.03 & 5.17 & 13.52 & 1203.875 \\ 
 & single & 12.14 & 0.37 & 9.8 & 16.96 & 1151.25 \\ \hline
\end{tabular}
\end{table*}

\begin{table*}[]
\caption{The similarity of reward signals for multiple agents based on multi-dimensional metrics. }
\begin{tabular}{ |p{1cm}||p{3cm}||p{1.5cm}|p{1.5cm}| p{1.5cm}|p{1.5cm}|p{3cm}|  }
 \hline
 \multicolumn{7}{|c|}{Similarity of reward signals} \\
 \hline
 Agents& Method &Mean&$\sigma$& $R_{min}$ & $R_{max}$ &  Training iterations\\
 \hline
2 & multi  & \textbf{0.925} & 0.039 & 0.424 & 0.961 & 7955.5   \\ 
 & hybrid & 0.898 & 0.045 & 0.741 & 0.951 & 9718.125 \\ 
 & single & 0.893 & 0.049 & 0.602 & 0.956 & 9276.625 \\ \hline
3 & multi  & 0.875 & 0.032 & 0.775 & 0.925 & 6514.125 \\ 
 & hybrid & \textbf{0.884} & 0.037 & 0.606 & 0.936 & 8259.25  \\ 
 & single & 0.846 & 0.025 & 0.73  & 0.891 & 7413.625 \\ \hline
4 & multi  & \textbf{0.864} & 0.031 & 0.7   & 0.908 & 5429.75  \\ 
 & hybrid & 0.844 & 0.02  & 0.789 & 0.99  & 6452.375 \\ 
 & single & 0.824 & 0.01  & 0.738 & 0.84  & 5338.125 \\ \hline
5 & multi  & \textbf{0.846} & 0.019 & 0.781 & 0.877 & 4609.375 \\ 
 & hybrid & 0.833 & 0.011 & 0.812 & 0.91  & 5200.75  \\ 
 & single & 0.821 & 0.009 & 0.716 & 0.835 & 4024.375 \\ \hline
6 & multi  & \textbf{0.832} & 0.013 & 0.716 & 0.854 & 3965.125 \\ 
 & hybrid & 0.826 & 0.007 & 0.814 & 0.91  & 4209.0   \\ 
 & single & 0.816 & 0.003 & 0.806 & 0.896 & 3124.25  \\ \hline
7 & multi  & \textbf{0.821} & 0.004 & 0.804 & 0.842 & 1753.75  \\ 
 & hybrid & 0.82  & 0.006 & 0.808 & 0.849 & 3284.125 \\ 
 & single & 0.812 & 0.003 & 0.802 & 0.875 & 2620.25  \\ \hline \hline
10 & multi & 0.81692 & 0.00299 & 0.80295 & 0.82205 & 1966.125 \\ 
 & hybrid & \textbf{0.81698} & 0.00313 & 0.79342 & 0.8242 & 1894.875 \\ 
 & single & 0.8154 & 0.00289 & 0.71693 & 0.82293 & 1695.5 \\ \hline
14 & multi & \textbf{0.81862} & 0.00272 & 0.767 & 0.83102 & 1212.625 \\ 
 & hybrid & 0.81736 & 0.00234 & 0.81467 & 0.85922 & 1203.875 \\ 
 & single & 0.81629 & 0.00217 & 0.78175 & 0.81996 & 1151.25 \\ \hline

\end{tabular}
\end{table*}
\subsection{Analysis of multiple reward signals}

Typically, reinforcement learning environments are considered as a black box that extracts a single reward signal, thus when all the agents return an individual reward, they are usually added and extracted as a scalar. Analysis of multiple reward signals should be undertaken and the following references have started addressing this aspect: 
\begin{itemize}
  \item Multi-objective reinforcement learning (MORL) with the need to find a Pareto optimal policy \cite{MORLGeneralized}, \cite{ScalarizedMORL}.
  \item Multi-objective optimization (MOO) is focused on non-conflicting objectives but suffers from a different pace of update. 
  \item Multi-Task reinforcement learning, where similarities in tasks are not recognized and the different pace of update indicates an ineffective use of the policy network \cite{MultiTaskLearning}, \cite{MultiTaskRL}.
  \item Multi-Agent reinforcement learning, where coordination of multiple agents is accomplished as part of a single actor policy network \cite{MultiAgentBook}, \cite{MultiAgentCooperation}, \cite{MultiAgentOverview} \cite{MultiAgentPartially}, \cite{MultiAgentRL}.
\end{itemize}

Multi-objective reinforcement learning deals with conflicting alternatives in a single environment \cite{Tzeng}. The action performed is synchronous with multiple objectives that extract reward signals of different values, unknown to the agent \cite{MORLGeneralized}, or based on a different position within the state-space but a constant penalty. The trade-off between competing objectives is hard to achieve using single-critic architectures that just consider an absolute reward. By extracting distinctive reward signals, an estimation of Pareto efficiency can be made. The naive approach of a classical weighted sum method \cite{WeightedSum} will fail for the case of a non-convex objective space. Epsilon-constraint based methods \cite{EpsilonConstraints} consider one main objective by applying hand-crafted restrictions for the sub-objectives. A generalization of the Bellman equation is used to learn a single parametric representation for optimal policies in \cite{MORLGeneralized}. One should mention that the classical scalar reward design to combine different objectives is likely to lead to diverse results as described in \cite{ProblemsAISafety}. 
In Multi-Task Reinforcement learning, the case of non-conflicting sub-objectives where multiple tasks have to be performed in distinct sub-environments, a single-critic network splits an action vector and feeds it to each sub-environment. The states are then stacked to a single state, but the reward signals remain distinct. Even when performing identical copies of the same task, the pace of learning is different and thus the actor policy does not make optimal use of already learned knowledge and fails to exploit similarities. Multiple tasks are competing in a sense for limited resources of a single learning system \cite{MultiTaskRL}, thus leading to emphasis on particular tasks depending on reward magnitude or density.

\subsection{Summary}

Single-Critic : 
\begin{equation}
C \to V(s)
\end{equation}
\begin{equation}
r \to C
\end{equation}
\begin{equation}
A(r, V(s), V(s'))
\end{equation}
\begin{equation}
\dim_0(A) = 1
\end{equation}
Multi-Critic :
\begin{equation}
C^1 \to V^1(s), C^2 \to V^2(s), ...
\end{equation}
\begin{equation}
r^i \to C^i
\end{equation}
\begin{equation}
A(r^1, r^2, ..., V^1(s), V^2(s), ..., V^1(s'), 
\end{equation}
\begin{equation}
\dim(a_0) \cdot \dim_0(A) = \dim_0(r(\theta))
\end{equation}
Hybrid-Critic : 
\begin{equation}
C^h \to v_1, v_2, v_3, ...
\end{equation}
\begin{equation}
r^i \to C^h
\end{equation}
\begin{equation}
A(r^1, r^2, .... v_1, v_2, v_3,...,... v_1', 
\end{equation}
\begin{equation}
\dim(a_0) \cdot \dim_0(A) = \dim_0(r(\theta))
\end{equation}

\subsection{Asynchronous approaches}
The approach of Asynchronous Advantage Actor-Critic (A3C) \cite{A2CA3C} is constructed having a global master network with a policy and value estimating network. Multiple workers are then initiated that interact with the copies of the environment. The workers are fed with global network weights and in turn, return their gradients to the global network. The workers are constructed to perform asynchronously and indeed some workers tend to be slightly out of date. The approach is engineered to utilize GPUs more efficiently and work better with large batch sizes \cite{policyGradientsLilianWeng}. The successor of A3C is similar but known as Advantage Actor-Critic (A2C). No own neural networks are needed for the workers, as they use the policy of the master network. The variability is achieved by each worker using its own environment. In this way, the same policy is applied in many parallel workers and the advantage is obtained from the mean reward. Regularization is used to encourage exploration where a side-objective in form of $H(\pi)$ is introduced: 
\begin{equation}
L = \mathbb{E}[A(s,a) \log \pi(s|a)] - H(\pi)
\end{equation}
with 
\begin{equation}
H(\pi) = - \sum_i \pi_i \log \pi_i
\end{equation}
The entropy of a distribution is maximized when the distribution is uniform.

\subsection{Classical Q-Learning}

First, we would like to introduce classical value-based reinforcement learning on the example of Q-Learning to give an enhanced understanding of the foundations of hybrid actor-critic approaches. The Q-table classically represents an action-value table for each state. A large number of states make it nearly impossible to simulate every combination of states and actions. For this reason, Q-Learning approximates this Q-function using neural networks. In deep Q-Learning architectures \cite{DeepQPartially, SelfLearningQ, QLearningPolicyGradient}, several actions are performed using the policy derived from the Q-table, and a target network is updated in an experienced replay. Deep Recurrent Q-Networks (DRQN) as discussed in \cite{DeepQPartially} and recurrent neural networks as presented in \cite{StanfordPaper} use Q-Learning to indirectly obtain the best policy. 
As a sub-form of temporal difference learning for a model free learning algorithm and for episodic and non-episodic tasks, the Q table of action-state is defined as:
\begin{equation}
Q(s) = Q(s,a) + \alpha(r + \gamma \max\limits_{a'} Q(s',a') - Q(s,a))
\end{equation}
In each iteration, the Q values for the actions are updated  using the reward and the temporal distance time, the discount rate $ \gamma \max\limits_{a'} Q(s',a') - Q(s,a)$. 
Since this is a regression problem, we can apply the format of regression Loss. 
\begin{equation}
L = (y_i - Q(s,a; \theta))^2
\end{equation}
with 
\begin{equation}
y_i = r + \gamma max_{a'} Q(s', a'; \theta)
\end{equation}
where we use two neural networks, the target network with frozen weights and the Q-network where training is performed. At the end of episode, the target network is reset to the Q-network. 

\subsection{Estimation of advantages}

In simple terms, the estimation of Advantages multiplied with the gradient of the logarithm of the policy distribution leads to: 
\begin{equation}
\nabla_\theta J(\theta) = \mathbb{E}[A(s,a)\nabla_{\theta} \log \pi(a|s)]
\end{equation}
The estimation can be carried out by
\begin{equation}
\nabla_\theta J(\theta) \approx \frac{1}{N} \sum_{i = 1}^N [A(s,a)\nabla_{\theta} \log \pi(a|s)]
\end{equation}
where the advantage is based on action-value per state minus the state-value:
\begin{equation}
A(s,a) = Q(s,a) - V(s)
\end{equation}
or 
\begin{equation}
A(s,a) = R + \gamma V'(s')- V(s)
\end{equation}
Q can be replaced by any N-step reward. The absolute reward is irrelevant if we know the advantage. No value estimation is accomplished by a second network parallel to the policy network $\pi_{\theta}$ which is $V_{\theta}$. The value estimation is typically trained on the Loss:
\begin{equation}L_v = \mathbb{E}[(G-V(s;\theta_v))^2] 
\end{equation} 
with 
\begin{equation}G= R + \gamma V(s')\end{equation}
or 
\begin{equation}G = R + \gamma R' + \gamma^2 R '' + ...\end{equation}


\end{document}